\documentclass{article}
\usepackage{iclr2027_conference,times}

\usepackage[utf8]{inputenc}
\usepackage[T1]{fontenc}
\usepackage{mathptmx}
\usepackage[hidelinks]{hyperref}
\usepackage{url}
\usepackage{booktabs}
\usepackage{amsmath}
\usepackage{amssymb}
\usepackage{graphicx}
\usepackage[table]{xcolor}
\usepackage{xspace}
\usepackage{subcaption}
\usepackage{multirow}
\usepackage{wrapfig}
\usepackage{needspace}

\usepackage{pgfplots}
\usepackage{pgfplotstable}
\usetikzlibrary{positioning, arrows.meta, calc, backgrounds, fit, patterns}
\pgfplotsset{compat=1.18}
\usepgfplotslibrary{groupplots}

\newcommand{\plotfont}{\fontfamily{ptm}\selectfont}
\tikzset{every picture/.append style={font=\plotfont}}

\definecolor{ours}{HTML}{FF4F2C}
\definecolor{ink}{HTML}{1A1A1A}
\definecolor{teach}{HTML}{3B1E1C}
\definecolor{clim}{HTML}{4A78B5}
\definecolor{geoc}{HTML}{3E9C7A}
\definecolor{satc}{HTML}{9A7AB0}
\definecolor{sinrc}{HTML}{C9973F}
\definecolor{cspc}{HTML}{B5476B}
\definecolor{gairc}{HTML}{6E8B3D}
\definecolor{ttec}{HTML}{17707E}
\definecolor{taxac}{HTML}{5B4B9E}
\definecolor{sledc}{HTML}{A8552A}
\definecolor{knn}{HTML}{6E6E6E}
\definecolor{protoR}{HTML}{0E7A8D}
\definecolor{protoS}{HTML}{A23B72}
\definecolor{soft}{HTML}{D8D3C6}

\pgfplotsset{
  tightaxis/.style={
    width=0.97\linewidth, height=1.55in,
    font=\plotfont,
    axis line style={ink, line width=0.5pt},
    axis x line*=bottom, axis y line*=left,
    tick align=outside, tick style={ink, line width=0.4pt},
    tick label style={font=\scriptsize, ink},
    label style={font=\small, ink},
    title style={font=\small\bfseries, ink, yshift=-1pt},
    legend cell align=left,
    legend style={font=\scriptsize, draw=none, fill=none, inner sep=1pt, row sep=-1.5pt},
    every axis plot/.append style={line width=0.9pt},
    clip mode=individual,
  },
  zeroline/.style={soft, line width=0.6pt, forget plot},
  faintgrid/.style={ymajorgrids, grid style={soft, line width=0.35pt}},
}

\pgfplotsset{
  ourmark/.style={mark=*, mark size=1.5pt, ours},
  basemark/.style={mark=o, mark size=1.4pt},
}

\tikzset{
  mind/flow/.style={->, >={Stealth[length=3pt]}, ink!70, line width=0.55pt},
  mind/skip/.style={->, >={Stealth[length=2.2pt]}, ink!55, line width=0.4pt},
  mind/box/.style={draw=ink!55, line width=0.6pt, rounded corners=2pt, fill=soft!20},
  mind/headsbox/.style={draw=ink!50, line width=0.5pt, rounded corners=2.5pt, fill=soft!16,
                        dash pattern=on 1pt off 1.1pt},
  mind/frozenbox/.style={draw=ink!45, line width=0.5pt, rounded corners=2.5pt, dashed},
  mind/tag/.style={font=\fontsize{5.6}{6.7}\selectfont, text=ink, fill=white, inner sep=1pt,
                   align=center},
  mind/cap/.style={font=\fontsize{5.6}{6.7}\selectfont, text=ink, inner sep=1pt, align=center},
  mind/label/.style={font=\fontsize{6.5}{7.8}\selectfont, text=ink},
  mind/opmath/.style={font=\fontsize{6.5}{7.8}\selectfont, text=ink, fill=white, inner sep=1pt},
  mind/grat/.style={ink!30, line width=0.3pt},
}

\DeclareMathAlphabet{\methodname}{OT1}{cmr}{m}{n}
\DeclareRobustCommand{\MIND}{\ensuremath{\methodname{MIND}}\xspace}
\DeclareRobustCommand{\MINDSET}{\ensuremath{\methodname{MINDSET}}\xspace}
\DeclareRobustCommand{\ChunkedPenalty}{\ensuremath{\methodname{Chunked~Penalty}}\xspace}
\DeclareRobustCommand{\MINDdim}[1]{\ensuremath{\methodname{MIND}_{#1}}\xspace}
\DeclareRobustCommand{\MINDFull}{\ensuremath{\methodname{MIND}_{\mathrm{Full}}}\xspace}
\newcommand{\FullDim}{\ensuremath{3{,}072}}
\DeclareRobustCommand{\MINDCP}{\ensuremath{\methodname{MIND{+}CP}}\xspace}
\definecolor{CiteGray}{HTML}{3F51B5}
\definecolor{FigPink}{HTML}{FF4081}

\hypersetup{
    colorlinks=true,
    citecolor=CiteGray, %
    linkcolor=FigPink,  %
    urlcolor=CiteGray   %
}

\newcommand{\doi}[1]{doi: \href{https://doi.org/#1}{\begingroup\urlstyle{rm}\nolinkurl{#1}\endgroup}}

\newcommand{\randwithin}{93}
\newcommand{\blockbeyond}{94}

\newcommand{\panelLeak}{\begin{tikzpicture}
\begin{axis}[tightaxis, faintgrid, xmode=log,
  xlabel={Distance to Nearest Training Point (km)}, ylabel={\% Within Distance},
  xmin=4, xmax=1200, ymin=0, ymax=104, ytick={0,25,50,75,100},
  xtick={5,10,25,50,100,250,500,1000}, xticklabels={5,10,25,50,100,250,500,1000},
  x tick label style={font=\scriptsize},
  legend style={at={(1.05,0.05)}, anchor=south east, font=\scriptsize, draw=none, fill=none,
    row sep=-3pt, inner sep=0.5pt}]
\addplot[protoR, mark=*, mark size=1.2pt] table[x=km, y=cum, col sep=comma] {figs/data/f2a-cdf-rand.csv};
\addlegendentry{Random Folds}
\addplot[protoS, mark=square*, mark size=1.1pt, dashed]
  table[x=km, y=cum, col sep=comma] {figs/data/f2a-cdf-block.csv};
\addlegendentry{$10^\circ$ Regions}
\draw[ink!35, line width=0.4pt] (axis cs:25,0) -- (axis cs:25,104);
\node[font=\scriptsize, protoR, anchor=north] at (axis cs:100,94) {\randwithin\% Within 25\,km};
\node[font=\scriptsize, protoS, anchor=south] at (axis cs:45,40) {$\blockbeyond\%$ Beyond 25\,km};
\end{axis}
\end{tikzpicture}}

\newcommand{\mindIconSat}[3]{\begin{scope}[shift={(#1,#2)}, rotate=38]
  \fill[#3!20] (-0.185,-0.05) rectangle (-0.075,0.05);
  \fill[#3!20] (0.075,-0.05) rectangle (0.185,0.05);
  \draw[#3, line width=0.4pt] (-0.185,-0.05) rectangle (-0.075,0.05);
  \draw[#3, line width=0.4pt] (0.075,-0.05) rectangle (0.185,0.05);
  \draw[#3, line width=0.4pt] (-0.13,-0.05) -- (-0.13,0.05) (0.13,-0.05) -- (0.13,0.05);
  \fill[#3] (-0.048,-0.048) rectangle (0.048,0.048);
\end{scope}}
\newcommand{\mindIconThermo}[3]{\begin{scope}[shift={(#1,#2)}]
  \draw[#3, line width=1.5pt, line cap=round] (0,0.14) -- (0,-0.04);
  \fill[#3] (0,-0.09) circle[radius=0.058];
  \draw[#3!50, line width=0.5pt] (0.05,0.105) -- (0.11,0.105) (0.05,0.03) -- (0.11,0.03);
\end{scope}}
\newcommand{\mindIconPhoto}[3]{\begin{scope}[shift={(#1,#2)}]
  \fill[#3!15] (-0.17,-0.12) rectangle (0.17,0.12);
  \fill[#3] (-0.06,0.05) circle[radius=0.028];
  \fill[#3!75] (-0.165,-0.115) -- (-0.05,0.01) -- (0.02,-0.05) -- (0.08,0.02)
               -- (0.165,-0.115) -- cycle;
  \draw[#3, line width=0.45pt] (-0.17,-0.12) rectangle (0.17,0.12);
\end{scope}}
\newcommand{\mindIconLeaf}[3]{\begin{scope}[shift={(#1,#2)}, rotate=-32]
  \fill[#3] (0,-0.11) .. controls (-0.11,-0.02) and (-0.085,0.08) .. (0,0.155)
            .. controls (0.085,0.08) and (0.11,-0.02) .. (0,-0.11);
  \draw[#3, line width=0.5pt, line cap=round] (0,-0.17) -- (0,-0.09);
  \draw[white, line width=0.4pt] (0,-0.06) -- (0,0.10);
\end{scope}}

\newcommand{\panelWidthCurves}[1][2.3in]{\begin{tikzpicture}
\begin{axis}[tightaxis, faintgrid, xmode=log, log basis x=2,
  xlabel={Truncation Dimension $m$}, ylabel={$R^2$}, height=#1, width=\linewidth,
  tick label style={font=\footnotesize, ink},
  label style={font=\small, ink},
  xtick={64,128,256,512,1024,2048,3072}, xticklabels={64,128,256,512,1k,2k,Full},
  xmin=56, xmax=3900, ymin=0.2, ymax=0.7, ytick={0.2,0.3,0.4,0.5,0.6,0.7}]
\addplot[zeroline, domain=56:3900, samples=2] {0};
\addplot[protoR, line width=1.1pt, mark=square*, mark size=1.8pt] table[x=width, y=macro, col sep=comma] {figs/data/f5a-random.csv};
\addplot[ours, dashed, line width=1.1pt, mark=*, mark options={solid}, mark size=1.8pt] table[x=width, y=macro, col sep=comma] {figs/data/f5a-spatial.csv};
\node[font=\small, protoR, anchor=west, inner sep=1pt] at (axis cs:130,0.64) {Random Folds};
\node[font=\small, ours, anchor=west, inner sep=1pt] at (axis cs:70,0.39) {$10^\circ$ Regional Holdout};
\end{axis}
\end{tikzpicture}}

\newcommand{\panelGapLadder}{\begin{tikzpicture}
\begin{axis}[tightaxis, faintgrid, xmode=log, height=2.3in,
  xlabel={Held-Out Block Size (degrees)}, ylabel={$R^2$},
  xtick={0.25,0.5,1,2,5,10,20,40}, xticklabels={0.25,0.5,1,2,5,10,20,40},
  xmin=0.22, xmax=45, ymin=-0.2, ymax=0.65, ytick={-0.2,0,0.2,0.4,0.6},
  legend style={at={(0.02,0.03)}, anchor=south west, font=\tiny, draw=none, fill=white,
    fill opacity=0.92, text opacity=1, row sep=-2pt, inner sep=1pt}]
\addplot[zeroline, domain=0.22:45, samples=2] {0};

  \addplot[ink!22, line width=0.5pt, forget plot]
    table[x=cell, y=macro, col sep=comma] {figs/data/hero-satclip.csv};

  \addplot[ink!22, line width=0.5pt, forget plot]
    table[x=cell, y=macro, col sep=comma] {figs/data/hero-sinr.csv};

  \addplot[ink!22, line width=0.5pt, forget plot]
    table[x=cell, y=macro, col sep=comma] {figs/data/hero-tte.csv};

  \addplot[ink!22, line width=0.5pt, forget plot]
    table[x=cell, y=macro, col sep=comma] {figs/data/hero-taxabind.csv};

  \addplot[ink!22, line width=0.5pt, forget plot]
    table[x=cell, y=macro, col sep=comma] {figs/data/hero-cspinat.csv};

  \addplot[ink!22, line width=0.5pt, forget plot]
    table[x=cell, y=macro, col sep=comma] {figs/data/hero-cspfmow.csv};

  \addplot[ink!22, line width=0.5pt, forget plot]
    table[x=cell, y=macro, col sep=comma] {figs/data/hero-gair.csv};

  \addplot[ink!22, line width=0.5pt, forget plot]
    table[x=cell, y=macro, col sep=comma] {figs/data/hero-sled.csv};
\addlegendimage{ink!22, line width=0.5pt}
\addlegendentry{Other INRs}
\addplot[ours, line width=1.0pt, mark=*, mark size=1.3pt]
  table[x=cell, y=macro, col sep=comma] {figs/data/hero-mindcp.csv};
\addlegendentry{\MINDCP}
\addplot[geoc, basemark]
  table[x=cell, y=macro, col sep=comma] {figs/data/hero-geoclip.csv};
\addlegendentry{GeoCLIP}
\addplot[clim, mark=square, mark size=1.4pt]
  table[x=cell, y=macro, col sep=comma] {figs/data/hero-climplicit.csv};
\addlegendentry{Climplicit}
\addplot[knn, dashed, mark=x, mark size=1.8pt]
  table[x=cell, y=macro, col sep=comma] {figs/data/hero-idw.csv};
\addlegendentry{Coordinate IDW}
\end{axis}
\end{tikzpicture}}

\newcommand{\panelVariogram}{\begin{tikzpicture}
\begin{axis}[tightaxis, xmode=log, height=1.75in,
  xlabel={Separation (km)},
  ylabel={Semivariance / Reference},
  xtick={10,100,1000,8000}, xticklabels={10,100,1000,8000},
  xmin=10, xmax=8000, ymin=0, ymax=2,
  ytick={0,0.5,1,1.5,2},
  legend columns=2, legend style={at={(0.03,0.97)}, anchor=north west, font=\tiny, draw=none,
    fill=white, fill opacity=0.85, text opacity=1, row sep=-3.2pt, column sep=2pt, inner sep=0.5pt}]
\addplot[soft, dashed, line width=0.6pt, forget plot] coordinates {(10,0.5) (8000,0.5)};
\node[font=\tiny, text=ink!60, anchor=south east] at (axis cs:8000,0.5) {Half Reference};
\addplot[clim] table[x=lag, y=n64, col sep=comma] {figs/data/f6c-vgram.csv};
\addlegendentry{$m=64$}
\addplot[ttec] table[x=lag, y=n128, col sep=comma] {figs/data/f6c-vgram.csv};
\addlegendentry{$m=128$}
\addplot[geoc] table[x=lag, y=n256, col sep=comma] {figs/data/f6c-vgram.csv};
\addlegendentry{$m=256$}
\addplot[gairc] table[x=lag, y=n512, col sep=comma] {figs/data/f6c-vgram.csv};
\addlegendentry{$m=512$}
\addplot[sinrc] table[x=lag, y=n1024, col sep=comma] {figs/data/f6c-vgram.csv};
\addlegendentry{$m=1$k}
\addplot[sledc] table[x=lag, y=n2048, col sep=comma] {figs/data/f6c-vgram.csv};
\addlegendentry{$m=2$k}
\addplot[ours, line width=1.1pt] table[x=lag, y=n3072, col sep=comma] {figs/data/f6c-vgram.csv};
\addlegendentry{\MINDFull}
\end{axis}
\end{tikzpicture}}

\newcommand{\panelHalfSill}{\begin{tikzpicture}
\begin{axis}[tightaxis, xmode=log, log basis x=2, ymode=log, height=1.75in,
  tick label style={font=\scriptsize, ink},
  xlabel={Truncation Dimension $m$},
  ylabel={Characteristic\\Distance (km)}, ylabel style={align=center},
  xtick={64,128,256,512,1024,2048,3072}, xticklabels={64,128,256,512,1k,2k,Full},
  xmin=56, xmax=3900, ymin=150, ymax=1000,
  ytick={150,300,600,1000}, yticklabels={150,300,600,1000}]
\addplot[ours, line width=1.1pt, mark=*, mark size=1.8pt] table[x=width, y=halfsill, col sep=comma] {figs/data/f6a-halfsill.csv};
\node[font=\small, ours, anchor=south west] at (axis cs:64,790) {790\,km};
\node[font=\small, ours, anchor=north east] at (axis cs:3072,196) {200\,km};
\end{axis}
\end{tikzpicture}}

\newcommand{\panelStorage}{\begin{tikzpicture}
\begin{axis}[tightaxis, faintgrid, xmode=log, log basis x=2, height=2.3in,
  xlabel={Representation Dimension}, ylabel={Random-Fold $R^2$},
  xmin=48, xmax=3900, ymin=0.4, ymax=0.675,
  xtick={64,128,256,512,1024,3072}, xticklabels={64,128,256,512,1k,3072},
  ytick={0.4,0.45,0.5,0.55,0.6,0.65}]
\addplot[ours, line width=1.0pt, mark=*, mark size=1.3pt]
  table[x=dims, y=random, col sep=comma] {figs/data/f8-mind.csv};
\node[font=\scriptsize, ours, anchor=south east] at (axis cs:2900,0.643) {\MIND};

\addplot[only marks, mark=square, mark size=1.5pt, clim] table[x=dims, y=random, col sep=comma] {figs/data/f8-climplicit.csv};
\node[font=\tiny, clim, anchor=west, xshift=3.5pt, inner sep=1pt] at (axis cs:1024,0.5891) {Climplicit};
\addplot[only marks, mark=o, mark size=1.6pt, geoc] table[x=dims, y=random, col sep=comma] {figs/data/f8-geoclip.csv};
\node[font=\tiny, geoc, anchor=north west, xshift=3.5pt, yshift=-2.5pt, inner sep=1pt] at (axis cs:512,0.5419) {GeoCLIP};
\addplot[only marks, mark=diamond, mark size=1.8pt, sinrc] table[x=dims, y=random, col sep=comma] {figs/data/f8-sinr.csv};
\node[font=\tiny, sinrc, anchor=east] at (axis cs:238,0.522) {SINR};
\addplot[only marks, mark=triangle, mark size=1.8pt, satc] table[x=dims, y=random, col sep=comma] {figs/data/f8-satclip.csv};
\node[font=\tiny, satc, anchor=north, yshift=-4.5pt, inner sep=1pt] at (axis cs:256,0.5044) {SatCLIP};

\addplot[only marks, mark=oplus, mark size=1.7pt, ttec] table[x=dims, y=random, col sep=comma] {figs/data/f8-tte.csv};
\node[font=\tiny, ttec, anchor=west] at (axis cs:548,0.5725) {TTE};
\addplot[only marks, mark=otimes, mark size=1.7pt, taxac] table[x=dims, y=random, col sep=comma] {figs/data/f8-taxabind.csv};
\node[font=\tiny, taxac, anchor=north west, xshift=3.5pt, yshift=-2.5pt, inner sep=1pt] at (axis cs:512,0.5115) {TaxaBind};
\addplot[only marks, mark=pentagon, mark size=1.7pt, cspc] table[x=dims, y=random, col sep=comma] {figs/data/f8-cspinat.csv};
\node[font=\tiny, cspc, anchor=south west, xshift=3.5pt, yshift=2.5pt, inner sep=1pt] at (axis cs:256,0.4288) {CSP-iNat};
\addplot[only marks, mark=pentagon*, mark size=1.4pt, cspc] table[x=dims, y=random, col sep=comma] {figs/data/f8-cspfmow.csv};
\node[font=\tiny, cspc, anchor=west, xshift=3pt] at (axis cs:256,0.4068) {CSP-fMoW};
\addplot[only marks, mark=asterisk, mark size=1.9pt, gairc] table[x=dims, y=random, col sep=comma] {figs/data/f8-gair.csv};
\node[font=\tiny, gairc, anchor=south west, xshift=3.5pt, yshift=2.5pt, inner sep=1pt] at (axis cs:768,0.4273) {GAIR};
\addplot[only marks, mark=halfcircle*, mark size=1.5pt, sledc] table[x=dims, y=random, col sep=comma] {figs/data/f8-sled.csv};
\node[font=\tiny, sledc, anchor=west, xshift=3.5pt, inner sep=1pt] at (axis cs:768,0.5566) {SLED};
\end{axis}
\end{tikzpicture}}

\newcommand{\mindChunkBar}[3]{\foreach \j/\offset/\chunkwidth in {1/0/0.12,2/0.14/0.12,3/0.28/0.17,4/0.47/0.23,5/0.72/0.33,6/1.07/0.46,7/1.55/0.46}{
    \pgfmathsetmacro{\left}{#1+\offset}
    \ifnum\j>#3
      \path[fill=ink!4, draw=ink!15, line width=0.35pt]
        (\left,#2) rectangle ++(\chunkwidth,0.20);
    \else
      \path[fill=ours!70, draw=ours!75!black, line width=0.35pt]
        (\left,#2) rectangle ++(\chunkwidth,0.20);
    \fi}}

\newcommand{\panelMethodOverview}{\resizebox{\linewidth}{!}{\begin{tikzpicture}[x=1cm, y=1cm, ink, font=\fontsize{7}{8.4}\selectfont,
  flow/.style={mind/flow},
  box/.style={draw=ink!45, fill=soft!15, rounded corners=2pt,
              line width=0.5pt, align=center, inner sep=4pt}]
\path[use as bounding box] (0,-3.27) rectangle (14,2.95);
\node[anchor=west, font=\fontsize{8}{9.6}\selectfont\bfseries] at (0,2.72) {(a) Nested Distillation};

\path[fill=clim!8, draw=ink!45, line width=0.5pt] (0.65,1.42) circle[radius=0.46];
\begin{scope}
  \clip (0.65,1.42) circle[radius=0.46];
  \foreach \rx in {0.18,0.35}{\draw[mind/grat] (0.65,1.42) ellipse[x radius=\rx, y radius=0.46];}
  \foreach \dy in {-0.3,-0.15,0,0.15,0.3}{\draw[mind/grat] (0.19,{1.42+\dy}) -- (1.11,{1.42+\dy});}
\end{scope}
\foreach \x/\y in {0.40/1.65,0.52/1.43,0.81/1.72,0.94/1.52,0.46/1.17,0.71/1.09,0.83/1.28}{
  \fill[ink!45] (\x,\y) circle[radius=0.02];}
\fill[ours] (0.98,1.40) circle[radius=0.038];
\draw[ours, line width=0.5pt] (0.98,1.40) circle[radius=0.077];
\node[align=center, anchor=north, font=\fontsize{6}{7.2}\selectfont] at (0.65,0.84) {Land Coords.\\$(\mathrm{lon},\mathrm{lat})$};
\draw[flow] (1.16,1.42) -- (1.64,1.42);
\path[box] (1.70,0.78) rectangle (3.12,2.06);
\draw[ours, line width=0.65pt] plot[domain=0:0.95, samples=40]
  ({1.90+\x}, {1.85+0.09*sin(\x*758)});
\foreach \y in {1.58,1.37,1.16}{
  \path[fill=soft!45, draw=ink!45, rounded corners=1pt, line width=0.4pt]
    (1.92,{\y-0.05}) rectangle (2.76,{\y+0.05});}
\foreach \ya/\yb in {1.70/1.46,1.49/1.25,1.28/1.04}{
  \draw[mind/skip] (2.81,\ya) to[out=-50,in=50] (2.81,\yb);}
\node[font=\fontsize{6.5}{7.8}\selectfont] at (2.41,0.94) {MIND};

\node[anchor=south] at (4.88,2.43) {Leading Chunks};
\node[anchor=south] at (7.63,2.43) {Linear Heads};
\foreach \y/\n/\dim in {2.06/1/64,1.29/4/512,0.52/7/3072}{\draw[flow] (3.12,1.42) to[out=0, in=180] (3.80,{\y+0.10});
  \mindChunkBar{3.86}{\y}{\n}
  \node[anchor=north] at (4.88,{\y-0.025})
    {$\mathbf{z}_{1:\ifnum\dim=3072 d\else\dim\fi}$};
  \draw[flow] (5.93,{\y+0.10}) -- (6.42,{\y+0.10});
  \path[fill=soft!12, draw=ink!35, rounded corners=2pt, line width=0.4pt]
    (6.48,{\y-0.05}) rectangle (8.68,{\y+0.29});
  \foreach \x/\c in {6.69/teach,7.21/clim,7.73/geoc,8.25/sinrc}{
    \path[fill=\c!45, draw=\c!80!black, line width=0.45pt]
      (\x,{\y-0.005}) -- (\x,{\y+0.245}) -- ({\x+0.21},{\y+0.19})
      -- ({\x+0.21},{\y+0.05}) -- cycle;}
  \node[anchor=north, font=\fontsize{6.5}{7.8}\selectfont] at (7.60,{\y+0.01})
    {\ifnum\dim=3072 $g_t$\else $g_t^{(\dim)}$\fi\ for Each Teacher $t$};
  \node (loss\dim) at (9.60,{\y+0.10}) {$\sum_t\ell$};
  \draw[flow] (8.73,{\y+0.10}) -- (loss\dim.west);
  \draw[flow] (10.54,{\y+0.10}) -- (loss\dim.east);
}

\path[mind/frozenbox] (10.59,0.28) rectangle (13.82,2.65);
\node[fill=white, inner sep=1.5pt] at (12.20,2.65) {Frozen Teacher Targets};
\foreach \y/\c/\nm/\sub in {2.16/teach/AEF/Satellite,
  1.64/clim/Climplicit/Climate,1.12/geoc/GeoCLIP/Photographs,0.60/sinrc/SINR/Species}{\path[fill=\c!5, draw=\c!65, line width=0.5pt, rounded corners=2pt]
    (10.75,{\y-0.22}) rectangle (13.67,{\y+0.22});
  \node[anchor=west, text=\c!75!black] at (11.35,\y) {\nm};
  \node[anchor=west, text=ink!65, font=\fontsize{6}{7.2}\selectfont]
    at (12.43,\y) {\sub};
}
\mindIconSat{11.07}{2.16}{teach}
\mindIconThermo{11.07}{1.64}{clim}
\mindIconPhoto{11.07}{1.12}{geoc}
\mindIconLeaf{11.07}{0.60}{sinrc}

\node[anchor=west, fill=white, inner sep=2pt, font=\fontsize{8}{9.6}\selectfont\bfseries]
  at (0,-0.43) {(b) Downstream Prediction};

\begin{scope}[yshift=-0.195cm]
\path[fill=clim!8, draw=ink!45, line width=0.5pt] (0.62,-1.50) circle[radius=0.40];
\begin{scope}
  \clip (0.62,-1.50) circle[radius=0.40];
  \foreach \rx in {0.16,0.30}{\draw[mind/grat] (0.62,-1.50) ellipse[x radius=\rx, y radius=0.40];}
  \foreach \dy in {-0.26,-0.13,0,0.13,0.26}{\draw[mind/grat] (0.22,{-1.50+\dy}) -- (1.02,{-1.50+\dy});}
\end{scope}
\foreach \x/\y in {0.42/-1.32,0.55/-1.62,0.78/-1.28,0.88/-1.55,0.66/-1.76,0.50/-1.48,0.74/-1.44}{
  \fill[ours] (\x,\y) circle[radius=0.028];}
\node[align=center, anchor=north, font=\fontsize{6}{7.2}\selectfont] at (0.62,-1.94) {Dataset\\Coords.};
\draw[flow] (1.06,-1.50) -- (1.46,-1.50);

\path[box, dashed] (1.52,-1.98) rectangle (2.74,-1.08);
\draw[ours, line width=0.55pt] plot[domain=0:0.82, samples=35]
  ({1.72+\x}, {-1.27+0.07*sin(\x*758)});
\foreach \y in {-1.45,-1.60}{\path[fill=soft!45, draw=ink!45, rounded corners=1pt, line width=0.35pt]
  (1.72,{\y-0.04}) rectangle (2.46,{\y+0.04});}
\draw[mind/skip] (2.52,-1.39) to[out=-50,in=50] (2.52,-1.55);
\node[font=\fontsize{6.5}{7.8}\selectfont] at (2.13,-1.82) {MIND};
\draw[flow] (2.74,-1.50) -- (3.12,-1.50);

\node[anchor=south, font=\fontsize{6.8}{7.8}\selectfont] at (4.19,-1.45) {Full Embedding $\mathbf{z}$};
\mindChunkBar{3.18}{-1.60}{7}
\end{scope}
\draw[flow] (5.19,-1.695) -- (5.35,-1.695) to[out=0, in=180] (5.95,-0.94);
\draw[flow] (5.19,-1.695) -- (5.35,-1.695) to[out=0, in=180] (5.95,-2.45);

\node[anchor=south, font=\fontsize{6.5}{7.8}\selectfont] at (6.96,-0.82) {Truncation};
\mindChunkBar{5.95}{-1.04}{3}
\draw[ink!60, dashed, line width=0.45pt] (6.41,-1.10) -- (6.41,-0.78);
\node[anchor=north, font=\fontsize{5.8}{7}\selectfont] (keptLabel) at (6.175,-1.08) {$\mathbf{z}_{1:m}$ Kept};
\node[anchor=north, font=\fontsize{5.8}{7}\selectfont, text=ink!55] (droppedLabel) at (7.195,-1.08) {Dropped};
\draw[flow] (7.96,-0.94) -- (8.36,-0.94);

\node[anchor=south, font=\fontsize{6.5}{7.8}\selectfont] (penaltyLabel) at (6.96,-2.40) {Chunked Penalty};
\node[fit=(keptLabel)(droppedLabel), inner sep=0pt, outer sep=0pt] (truncationLabels) {};
\coordinate (orMidpoint) at ($(truncationLabels.south)!0.5!(penaltyLabel.north)$);
\node[font=\fontsize{7}{8.4}\selectfont\bfseries] at (6.96,0 |- orMidpoint) {OR};
\mindChunkBar{5.95}{-2.55}{7}
\foreach \j/\offset/\chunkwidth in {1/0/0.12,2/0.14/0.12,3/0.28/0.17,4/0.47/0.23,5/0.72/0.33,6/1.07/0.46,7/1.55/0.46}{
  \path[fill=ours!30, draw=ours!70, line width=0.3pt]
    ({5.95+\offset},-2.91) rectangle ++(\chunkwidth,{0.04*\j});
}
\node[anchor=north, font=\fontsize{5.8}{7}\selectfont, text=ink!65] at (6.96,-2.95)
  {Penalty Weight $\alpha\gamma^{c(i)}$};
\draw[flow] (7.96,-2.45) -- (8.36,-2.45);

\node[box, fill=white, minimum width=1.45cm, minimum height=0.44cm]
  (ridgeSelect) at (9.10,-0.94) {Ridge};
\node[box, fill=white, minimum width=1.45cm, minimum height=0.44cm]
  (ridgePenalty) at (9.10,-2.45) {Ridge + CP};
\node[draw=ink!45, fill=white, rounded corners=2pt, line width=0.45pt, inner sep=2pt,
  font=\fontsize{6}{7.2}\selectfont] (targets) at (9.10,-1.695) {Targets $y$};
\draw[mind/skip] (targets.north) -- (ridgeSelect.south);
\draw[mind/skip] (targets.south) -- (ridgePenalty.north);
\node[box, fill=ours!8, draw=ours!55, minimum width=1.30cm, minimum height=0.70cm]
  (predictionSelect) at (10.90,-0.94) {Prediction\\$\hat{y}_{\mathrm{trunc}}$};
\node[box, fill=ours!8, draw=ours!55, minimum width=1.30cm, minimum height=0.70cm]
  (predictionPenalty) at (10.90,-2.45) {Prediction\\$\hat{y}_{\mathrm{CP}}$};
\draw[flow] (ridgeSelect.east) -- (predictionSelect.west);
\draw[flow] (ridgePenalty.east) -- (predictionPenalty.west);

\path[fill=clim!8, draw=ink!45, line width=0.5pt] (12.90,-1.695) circle[radius=0.44];
\begin{scope}
  \clip (12.90,-1.695) circle[radius=0.44];
  \foreach \rx in {0.18,0.34}{\draw[mind/grat] (12.90,-1.695) ellipse[x radius=\rx, y radius=0.44];}
  \foreach \dy in {-0.28,-0.14,0,0.14,0.28}{\draw[mind/grat] (12.46,{-1.695+\dy}) -- (13.34,{-1.695+\dy});}
  \foreach \x/\y in {0.42/-1.32,0.55/-1.62,0.78/-1.28,0.88/-1.55,0.66/-1.76,0.50/-1.48,0.74/-1.44}{
    \fill[ours] ({12.90+1.1*(\x-0.62)},{-1.695+1.1*(\y+1.50)}) circle[radius=0.031];}
  \foreach \dx/\dy in {0/0.34,-0.30/-0.20,0.28/-0.25}{
    \fill[geoc] ({12.90+1.1*\dx},{-1.695+1.1*\dy}) circle[radius=0.034];}
\end{scope}
\draw[flow] (predictionSelect.east) to[out=0,in=140] (12.46,-1.695);
\draw[flow] (predictionPenalty.east) to[out=0,in=-140] (12.46,-1.695);
\node[anchor=north, font=\fontsize{6}{7.2}\selectfont] at (12.90,-2.21) {Predicted Coords.};
\end{tikzpicture}}}

\title{MIND the Gap: {\Large A Geographic Implicit Neural\\\ Representation with Adjustable Spatial Scale}}
\iclrfinalcopy
\title{MIND the Gap: {\fontsize{14}{17}\selectfont A Geographic Implicit Neural\\Representation with Adjustable Spatial Scale}\par\vspace{12pt}}
\author{%
  \parbox{\dimexpr\textwidth-2\tabcolsep\relax}{\centering
    \normalfont\fontsize{9}{11}\selectfont
    {\hypersetup{urlcolor=black}%
    \href{https://isaac.earth}{\textbf{Isaac Corley}}\textsuperscript{1}\thanks{Corresponding author: \href{mailto:isaac.corley@taylorgeospatial.org}{\texttt{isaac.corley@taylorgeospatial.org}}.}\quad
    \href{https://arjunashokrao.me/}{\textbf{Arjun Rao}}\textsuperscript{2,5}\quad
    \href{https://www.estherrolf.com/}{\textbf{Esther Rolf}}\textsuperscript{3}\quad
    \href{https://konstantinklemmer.github.io/}{\textbf{Konstantin Klemmer}}\textsuperscript{4}\quad
    \href{https://imaginarynumber.net/}{\textbf{Evan Shelhamer}}\textsuperscript{2,5}\\[2pt]
    \href{https://nilsleh.github.io/}{\textbf{Nils Lehmann}}\textsuperscript{6}\quad
    \href{https://marcrusswurm.com/}{\textbf{Marc Ru\ss wurm}}\textsuperscript{7}\quad
    \href{https://gengchenmai.github.io/}{\textbf{Gengchen Mai}}\textsuperscript{8}\quad
    \href{https://jacobsn.github.io/}{\textbf{Nathan Jacobs}}\textsuperscript{9}\quad
    \href{https://hannah-rae.github.io/}{\textbf{Hannah Kerner}}\textsuperscript{1,10}\\[5pt]}
    \fontsize{8}{9.5}\selectfont
    \textsuperscript{1}Taylor Geospatial\quad
    \textsuperscript{2}University of British Columbia\quad
    \textsuperscript{3}University of Colorado Boulder\\
    \textsuperscript{4}University College London\quad
    \textsuperscript{5}Vector Institute\quad
    \textsuperscript{6}Technical University of Munich\quad
    \textsuperscript{7}University of Bonn\\
    \textsuperscript{8}University of Texas at Austin\quad
    \textsuperscript{9}Washington University in Saint Louis\quad
    \textsuperscript{10}Arizona State University\\[10pt]
    {\fontsize{9}{11}\selectfont\bfseries\hypersetup{urlcolor=ours}\href{https://research.taylorgeospatial.org/mind/}{research.taylorgeospatial.org/mind}}
  }%
}
\AddToHook{env/abstract/before}{\vspace{-12pt}}
\hypersetup{
  pdftitle={MIND the Gap: A Geographic Implicit Neural Representation with Adjustable Spatial Scale},
  pdfauthor={Isaac Corley, Arjun Rao, Esther Rolf, Konstantin Klemmer, Evan Shelhamer, Nils Lehmann, Marc Ru\ss wurm, Gengchen Mai, Nathan Jacobs, Hannah Kerner},
  pdfsubject={}
}

\date{}

\begin{document}
\maketitle
\fancyhead{}
\renewcommand{\headrulewidth}{0pt}

\begin{abstract}
	Geographic measurements are often sparse, leaving large areas without labels for the quantities we want to map. Geographic implicit neural representations (INRs) provide coordinate-based embeddings that can be combined with sparse labels to predict at unsampled locations without satellite imagery at inference. Yet existing INRs are largely evaluated with random holdouts, leaving their ability to generalize across larger geographic gaps unclear. We introduce Matryoshka Implicit Neural Distillation (\MIND), a geographic INR whose spatial granularity can be adjusted after training. \MIND distills several pretrained geospatial models using nested supervision at increasing embedding dimensions, dividing the representation into contiguous chunks. Early chunks capture broad spatial patterns, while later chunks add increasingly local variation. Downstream models can retain only the leading chunks or use our \ChunkedPenalty to reduce reliance on later chunks without retraining the INR. We also introduce CoordBench, comprising $52$ datasets and $78$ targets with both random and regional holdouts at multiple spatial scales. Across CoordBench, fine-scale features help most when labels are nearby, while smoother representations generalize better across larger geographic gaps. \MIND with the \ChunkedPenalty achieves the highest aggregate regression and classification performance among tested INRs and the highest overall performance under regional holdout. These results show that geographic representations should be evaluated and adapted according to the spatial separation between labeled and prediction locations.
\end{abstract}

\section{Introduction}
Measurements of income, health, and other socioeconomic and environmental quantities are unevenly distributed across the world, while maps require dense predictions between and beyond these observations for visualization and decision-making~\citep{mosaiks2021,pdfm2024,cdcplaces2023,sustainbench2021,worldclim2017}. Some unobserved locations lie near labeled samples, while others lie in regions with few or no labels for the target of interest~\citep{mai2020space2vec}. Geographic implicit neural representations (INRs) encode satellite observations and other geospatial data as a continuous function over the spherical surface of planet Earth~\citep{mai2022review,mai2020space2vec,russwurm2023}. Queried via location (latitude/longitude) inputs, these implicit data representations can be combined with available labels to interpolate or extrapolate at new unsampled locations, without requiring satellite imagery at inference time~\citep{geoclip2023,satclip2024}.

\begin{figure}[!t]
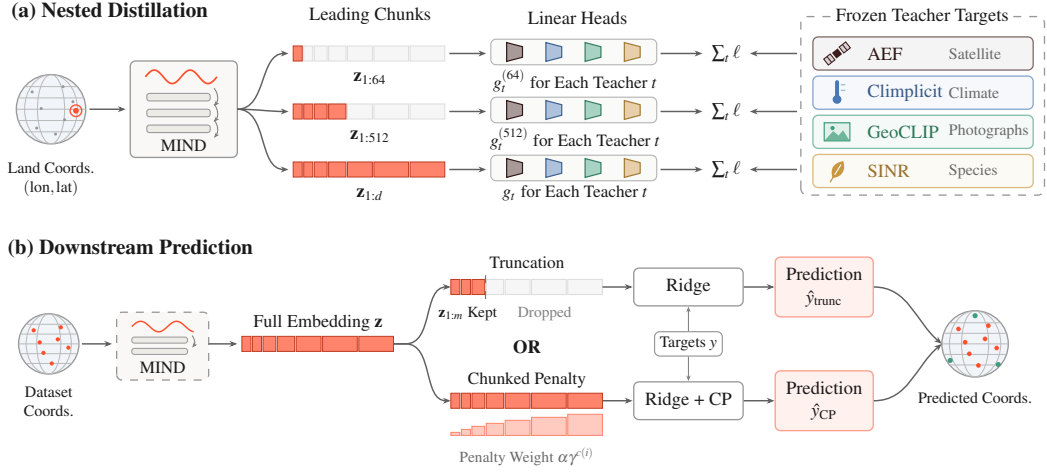

	\centering
	\panelMethodOverview
	\caption{\textbf{MIND training and downstream prediction.}
		(a) Training learns a shared embedding of multiple teachers. At each supervised dimension, linear heads reconstruct all teacher embeddings from the leading chunks. The heads are discarded after training.
		(b) Prediction takes the full embedding, truncates (keeping leading chunks), or regularizes it (penalizing later chunks) to suit the needs of the task.}
	\label{fig:method}
\end{figure}

Standard evaluations of INRs fit linear predictors to frozen features and evaluate on randomly assigned folds of the labeled dataset~\citep{climplicit2025,unigeoclip2026,gair2025,csp2023}. Because nearby locations often have similar target values due to \emph{spatial autocorrelation}, these splits primarily test local interpolation near labeled locations~\citep{ploton2020spatial,shen2026blisco}. In our benchmark, $\randwithin\%$ of random-fold test observations lie within $25$\,km of a training point, and inverse-distance weighting (IDW) on coordinates achieves higher aggregate regression $R^2$ than every tested INR under random folds (Table~\ref{tab:coordbench}). Under regional holdout, features that capture local variation can lose predictive accuracy as held-out regions grow, and the ranking of encoders changes. Evaluating across these scales is necessary when maps extend beyond surveyed areas.

A general-purpose coordinate representation should combine diverse geographic signals while supporting the different spatial scales needed for nearby and regional prediction. Existing INRs are often trained and evaluated within one modality, such as climate, photographs, or species observations~\citep{climplicit2025,geoclip2023,sinr2023}. Distilling specialist embeddings at shared coordinates offers a way to combine these signals without processing the original datasets.

We introduce \MIND (Matryoshka Implicit Neural Distillation), a geographic INR with adjustable spatial scale (Fig.~\ref{fig:method}). \MIND learns to reconstruct embeddings from pretrained models, called \emph{teachers}, using \MINDSET, our dataset of paired coordinates and teacher embeddings. Following Matryoshka Representation Learning~\citep{mrl2022}, we supervise cumulative sets of dimensions at $64$, $128$, and successively larger sizes up to the full embedding. These boundaries define contiguous \emph{chunks}, with each reconstruction using all preceding chunks. In our experiments, early chunks capture smoother, large-scale geographic patterns, while later chunks add more localized variation. The encoder produces a fixed embedding at each coordinate, with spatial scale selected through truncation or coefficient regularization when fitting the downstream predictor. Our \emph{\ChunkedPenalty} (CP) retains all chunks but penalizes later chunks more strongly, balancing performance under random folds and regional holdout without retraining the encoder.

\begin{figure}[!t]
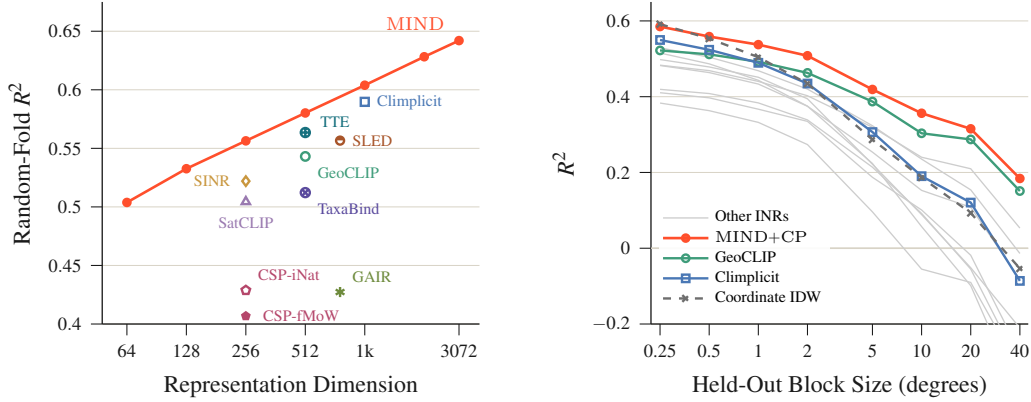

	\centering
	\begin{subfigure}{0.485\linewidth}\panelStorage\end{subfigure}\hfill
	\begin{subfigure}{0.485\linewidth}\panelGapLadder\end{subfigure}
	\caption{\textbf{Representation dimension and geographic separation.} Without separation, \MIND has higher random-fold $R^2$ than the other tested INRs (left). With enforced geographic separation of the train and test sets, \MIND with the \ChunkedPenalty (\MINDCP) has the highest $R^2$ among tested INRs at each spatial scale (right), exceeding IDW at block sizes of $0.5^\circ$ and larger.}
	\label{fig:hero}
\end{figure}

Prior geographic INR studies use different datasets and evaluation procedures, making encoder comparisons difficult. To enable consistent comparisons, we introduce CoordBench, with $52$ datasets and $78$ targets evaluated using common predictors, metrics, random folds, and regional holdout. \MINDCP achieves the highest aggregate scores among the tested INRs (Fig.~\ref{fig:hero}, right) and improves regional regression over selecting a fixed number of leading chunks to retain. The same \MIND encoder trained on the same data without nested supervision does not show the coarse-to-fine ordering across chunks, and the benefit of the penalty depends on the trained dimension order. We release \MIND, CoordBench, \MINDSET, and a global $1$\,km grid of \MIND embeddings, so users can select chunks without recomputing embeddings.

\section{Related work}
\paragraph{Geographic representations.}
Geographic INRs encode coordinates using multiscale features, spherical harmonics, or learned networks~\citep{mai2020space2vec,sphere2vec2023,russwurm2023,unigeoclip2026}. Sphere2Vec explicitly encodes spherical geometry to preserve geodesic distance~\citep{sphere2vec2023}. Their supervision includes imagery~\citep{satclip2024,geoclip2023,csp2023,gair2025}, species observations~\citep{sinr2023,taxabind2025}, climate data~\citep{climplicit2025}, and geographic partitions~\citep{cher2026tte}. Other Earth embeddings include AlphaEarth Foundations (AEF)~\citep{alphaearth2025} and models derived from multiple geospatial data sources~\citep{pdfm2024,earthembeddings2026}. Scalable Location Encoding via Distillation (SLED)~\citep{sled2026}, the closest distillation approach to \MIND, distills frozen satellite image encoders into a coordinate representation. \MIND distills teachers trained on satellite imagery, climate data, photographs, and species observations into a single coordinate embedding.

\paragraph{Nested supervision and spatial scale.}
Matryoshka Representation Learning supervises cumulative sets of dimensions so one embedding can serve different dimension budgets~\citep{mrl2022,smec2025}. TESSERA v2 uses Matryoshka distillation to produce compact image-derived embeddings at several dimensions~\citep{feng2026tesserav2scalingpixelwise}. Neural fields can explicitly constrain spatial frequencies~\citep{bacon2022,barf2021,freenerf2023}. RANGE~\citep{dhakal2025range} augments frozen SatCLIP embeddings through an external image-feature bank and controls spatial smoothness through retrieval weights. \MIND learns a coordinate embedding with nested supervision and adjusts its spatial scale through downstream truncation or the \ChunkedPenalty.

\paragraph{Geographic INR evaluation.}
Regional holdout is an established practice to address spatial autocorrelation during evaluation in ecology and remote sensing~\citep{roberts2017crossvalidation,valavi2019blockcv,karasiak2022spatial,rolf2023evaluation}. Spatial baselines and analyses of train--test distances show that geospatial evaluations need to account for proximity to training data~\citep{hijmans2012crossvalidation,bahn2007niche}. Buffered exclusion removes nearby training points directly~\citep{shen2024bloo,shen2026blisco}, and distance-matched folds aim to reproduce the expected distances between train and test locations~\citep{mila2022nndm}. \citet{cai2025no} introduce spherical wavelet encodings and the FAIR-Earth dataset for evaluating geographic subgroup disparities. CoordBench complements these and other INR evaluations~\citep{climplicit2025,unigeoclip2026,gair2025,satclip2024,torchspatial2024} by testing both nearby prediction and generalization across larger geographic gaps.

\section{\MIND for Geographic Representations}
\label{sec:method}
Matryoshka Implicit Neural Distillation (\MIND) maps a geographic coordinate (latitude/longitude) to an embedding divided into contiguous chunks. In our experiments, earlier chunks capture broader spatial variation than later chunks (Sec.~\ref{sec:width}). During training, linear layers called reconstruction heads map each cumulative set of chunks to the teacher embeddings (Fig.~\ref{fig:method}). After training, we discard these heads and fit a downstream predictor to either the leading chunks or the full embedding.

\begin{table*}[t!]
	\centering\scriptsize
	\setlength{\tabcolsep}{3pt}
	\renewcommand{\arraystretch}{1.15}
	\caption{\textbf{CoordBench sources and coverage.} CoordBench is composed of 78 target variables across 52 datasets, combining the disjoint benchmark datasets used in many Geographic INR evaluations. Spatial unit gives the source label unit or raster resolution where available.\label{tab:bench}}
	\begin{tabular*}{\textwidth}{@{\extracolsep{\fill}}>{\raggedright\arraybackslash}p{0.34\textwidth}>{\centering\arraybackslash}p{0.26\textwidth}>{\centering\arraybackslash}p{0.23\textwidth}>{\centering\arraybackslash}p{0.12\textwidth}@{}}
		\toprule
		Source             & Targets                       & Spatial unit          & Coverage           \\
		\midrule
		PDFM~\citep{pdfm2024} & $27$ socioeconomic measures    & US place-level         & Contiguous US \\
		PLACES~\citep{cdcplaces2023} & $12$ health measures           & ZCTA centroids         & US postal areas \\
		SustainBench~\citep{sustainbench2021} & Assets, water, sanitation      & DHS cluster centroids  & Global \\
		MOSAIKS/USAVars~\citep{mosaiks2021} & Population, nightlights, income & $1$\,km points       & US \\
		SatCLIP~\citep{satclip2024} & Population, elevation, biome  & Global point samples   & Global \\
		AEF evaluation~\citep{alphaearth2025} & Land cover, crops, emissivity & Task-dependent         & Global \\
		WorldClim~\citep{worldclim2017} & Temperature, precipitation    & $10$-arc-minute grid  & Land \\
		SoilGrids~\citep{soilgrids2021} & Organic carbon, pH            & Global raster points  & Land \\
		California Housing~\citep{pace1997sparse} & Median house value            & Census block groups   & California \\
		\bottomrule
	\end{tabular*}
\end{table*}

\begin{figure}[!t]
	\begin{minipage}[t]{0.485\linewidth}
		\vspace{0pt}
		\centering
		\panelWidthCurves[1.5in]
		\caption{\textbf{Chunk performance across different holdouts.} Adding chunks improves random-fold $R^2$ but lowers $10^\circ$ blocked regional-holdout $R^2$ beyond $128$ dimensions.}
		\label{fig:dims}
	\end{minipage}\hfill
	\begin{minipage}[t]{0.485\linewidth}
		\vspace{0pt}
		\captionsetup{type=table}
		\centering
\caption{\textbf{The effect of nested supervision.} At the same dimension, \MIND's leading chunks have higher $R^2$ on CoordBench than the identical model trained without nested supervision, with larger differences under regional holdout.\label{tab:fixw-main}}
\begingroup
\small
	\setlength{\tabcolsep}{3pt}
	\renewcommand{\arraystretch}{1.15}
	\begin{tabular}{@{}llccc@{}}
		\toprule
		Dim. & Model & Random & $10^\circ$ & $40^\circ$ \\
		\midrule
		\multirow{2}{*}{$64$} & Non-nested & 0.452 & 0.225 & $-0.019$ \\
		& \MIND & \textbf{0.504} & \textbf{0.309} & \textbf{0.128} \\
		\midrule
		\multirow{2}{*}{$256$} & Non-nested & 0.550 & 0.312 & 0.059 \\
		& \MIND & \textbf{0.556} & \textbf{0.322} & \textbf{0.116} \\
		\bottomrule
	\end{tabular}
	\par
\endgroup

	\end{minipage}
\end{figure}

\subsection{Nested distillation}
\label{sec:pretrain}
Following the notation of Matryoshka Representation Learning~\citep{mrl2022}, let $\mathbf{z}=f_\theta(\mathbf{x})\in\mathbb{R}^{d}$ be the full embedding at coordinate $\mathbf{x}$. Truncation at $m$ keeps the first $m$ dimensions, which form $\mathbf{z}_{1:m}\in\mathbb{R}^{m}$, where $m\in[d]$ and $[d]=\{1,\ldots,d\}$. We denote this representation by \MINDdim{m}. The supervised dimensions form a set $\mathcal{M}\subset[d]$ that includes $d$, with consecutive values defining the chunk boundaries. Each teacher provides a target vector $\mathbf{e}_t\in\mathbb{R}^{d_t}$ at the same coordinate. We normalize the teacher targets by the PHI-S transformation~\citep{phis2024} and minimize the following distillation loss:

\begin{equation}
	\mathcal{L} = \sum_t \ell\bigl(g_t(\mathbf{z}),\mathbf{e}_t\bigr)
	+ \frac{1}{|\mathcal{M}|-1}\sum_{m\in\mathcal{M}\setminus\{d\}}\sum_t
	\ell\bigl(g_t^{(m)}(\mathbf{z}_{1:m}),\mathbf{e}_t\bigr).
	\label{eq:mind}
\end{equation}

Each teacher has a separate linear reconstruction head at every supervised dimension. The head $g_t^{(m)}$ maps $m$ student dimensions to all $d_t$ teacher dimensions, while $g_t$ uses the full $d$-dimensional student embedding. For teacher $t$, let $\mathbf{u}$ and $\mathbf{v}$ be the $d_t$-dimensional prediction and target. The reconstruction loss combines cosine distance and mean squared error:

\begin{equation}
	\ell(\mathbf{u},\mathbf{v}) = 1-\bar{\mathbf{u}}^\top\bar{\mathbf{v}}
	+\frac{1}{d_t}\lVert\bar{\mathbf{u}}-\bar{\mathbf{v}}\rVert_2^2,
	\qquad \bar{\mathbf{u}}=\frac{\mathbf{u}}{\lVert\mathbf{u}\rVert_2}.
	\label{eq:reconstruction}
\end{equation}

Here $\bar{\mathbf{v}}=\mathbf{v}/\lVert\mathbf{v}\rVert_2$. App.~\ref{app:repro} gives the PHI-S normalization details. We weight the full-dimensional loss separately and average the losses for $m<d$~\citep{smec2025}. We measure how spatial scale changes as more chunks are included in Sec.~\ref{sec:width}.

Our encoder is a residual sinusoidal representation network (ReSIREN)~\citep{climplicit2025,siren2020} that maps coordinates from the Equal Earth map projection to an embedding of full dimension $d=\FullDim$, denoted \MINDFull. The teachers are AEF, Climplicit, GeoCLIP, and Spatial Implicit Neural Representations (SINR), which are supervised by satellite imagery, gridded climate data, photographs, and species observations, respectively~\citep{alphaearth2025,climplicit2025,geoclip2023,sinr2023}. We train on \MINDSET, a pretraining dataset we compiled that pairs teacher embeddings with $12$M land coordinates sampled densely in and around cities (App.~\ref{app:mindset}). Architecture, teacher dimensions, and optimization settings are in App.~\ref{app:repro}.

\subsection{\ChunkedPenalty for Scale-aware Regularization}
\label{sec:readtime}
Truncation restricts the predictor to the leading chunks, discarding later dimensions and any useful information they contain. As a soft alternative, we propose the \ChunkedPenalty to regularize the full embedding by penalizing coefficients for later chunks more strongly. For features $X$ and regression targets $y$, the predictor coefficients solve

\begin{table*}[t]\centering\small
	\setlength{\tabcolsep}{4pt}
	\caption{\textbf{Performance of Different Methods on CoordBench.} \MINDCP has the highest $R^2$ and accuracy among tested INRs while Coordinate IDW has the highest random-fold $R^2$. Degrees indicate the block size for regional holdout, and we report means across five fold-assignment seeds. Appendix Table~\ref{tab:coordbench-spread} reports standard deviations. Best result in \textbf{bold}, second-best in \textit{italics}.}\label{tab:coordbench}
	\begin{tabular}{@{}lcccccccccc@{}}
		\toprule
		& \multicolumn{5}{c}{$R^2$} & \multicolumn{5}{c}{Accuracy (\%)} \\
		\cmidrule(lr){2-6}\cmidrule(l){7-11}
		Method & Random & $2^\circ$ & $10^\circ$ & $20^\circ$ & $40^\circ$ & Random & $2^\circ$ & $10^\circ$ & $20^\circ$ & $40^\circ$ \\
		\midrule
		Coord. IDW & \textbf{0.675} & 0.435 & 0.187 & 0.093 & -0.053 & 67.7 & 62.8 & 55.7 & 53.0 & 51.2 \\
		\midrule
		Cartesian 3D & 0.137 & 0.089 & -0.088 & -0.173 & -0.423 & 51.8 & 50.6 & 46.4 & 45.8 & 44.4 \\
		Wrap (Sin/Cos) & 0.204 & 0.155 & -0.025 & -0.116 & -0.344 & 53.1 & 52.2 & 48.3 & 46.8 & 45.2 \\
		\midrule
		SINR \textit{(ICML '23)} & 0.522 & 0.394 & 0.089 & -0.051 & -0.211 & 63.8 & 61.4 & 55.2 & 51.4 & 49.5 \\
		SatCLIP \textit{(AAAI '25)} & 0.504 & 0.375 & 0.153 & 0.108 & -0.082 & 63.8 & 60.8 & 55.0 & 52.4 & 51.9 \\
		Climplicit \textit{(ICLRW '25)} & 0.590 & 0.435 & 0.190 & 0.120 & -0.086 & 66.7 & 63.2 & 57.1 & 54.6 & 52.8 \\
		GeoCLIP \textit{(NeurIPS '23)} & 0.543 & 0.463 & 0.303 & 0.287 & \textit{0.151} & 63.5 & 60.6 & 56.0 & 54.6 & 54.2 \\
		CSP-iNat \textit{(ICML '23)} & 0.429 & 0.339 & 0.092 & -0.055 & -0.310 & 61.4 & 59.4 & 54.0 & 51.3 & 48.2 \\
		CSP-fMoW \textit{(ICML '23)} & 0.407 & 0.335 & 0.101 & -0.018 & -0.276 & 61.7 & 60.0 & 54.8 & 51.9 & 47.7 \\
		GAIR \textit{(ISPRS '26)} & 0.427 & 0.274 & -0.055 & -0.090 & -0.322 & 60.3 & 57.1 & 51.7 & 49.6 & 48.5 \\
		TTE \textit{(ECCV '26)} & 0.563 & 0.420 & 0.235 & 0.154 & -0.014 & 66.2 & 62.5 & 58.2 & 55.8 & 54.7 \\
		TaxaBind \textit{(WACV '25)} & 0.512 & 0.402 & 0.240 & 0.210 & 0.053 & 63.9 & 61.0 & 55.9 & 54.4 & 53.7 \\
		SLED \textit{(arXiv '26)} & 0.557 & 0.375 & 0.057 & -0.099 & -0.394 & 66.0 & 62.0 & 53.9 & 49.9 & 46.0 \\
		\midrule
		\MINDdim{64} & 0.504 & 0.447 & 0.309 & 0.281 & 0.128 & 62.5 & 60.5 & 56.9 & 55.3 & 54.2 \\
		\MINDdim{128} & 0.533 & 0.465 & \textit{0.327} & \textit{0.303} & 0.136 & 64.0 & 62.0 & 57.7 & 56.3 & 54.8 \\
		\MINDdim{256} & 0.556 & 0.476 & 0.322 & 0.279 & 0.116 & 65.2 & 63.0 & 58.1 & \textit{56.8} & \textit{55.3} \\
		\MINDFull & 0.642 & \textit{0.492} & 0.280 & 0.211 & 0.019 & \textit{67.8} & \textit{64.4} & \textit{59.1} & 56.2 & 54.6 \\
		\rowcolor{ours!15}
		\MINDCP & \textit{0.643} & \textbf{0.508} & \textbf{0.356} & \textbf{0.315} & \textbf{0.184} & \textbf{67.8} & \textbf{64.6} & \textbf{59.9} & \textbf{57.8} & \textbf{56.3} \\
		\bottomrule
	\end{tabular}
\end{table*}

\begin{equation}
	(\hat{\beta},\hat{b})=\arg\min_{\beta,b} \lVert y-X\beta-b\mathbf{1}\rVert_2^2
	+\alpha\sum_i\gamma^{c(i)}\beta_i^2,
	\label{eq:penalty}
\end{equation}

Here $\beta$ contains the feature coefficients and $b$ is the intercept. The chunk index $c(i)$ is fixed by the training boundaries, with $c(i)=0$ for dimensions $1$--$64$, $c(i)=1$ for $65$--$128$, and so on. The parameter $\alpha$ sets the overall penalty and $\gamma$ determines how much it grows between chunks. Standard ridge regression is the special case $\gamma=1$. \MINDCP applies the \ChunkedPenalty to the same full embedding used by \MINDFull with standard ridge regression. The penalty downweights later chunks, where we observe more local spatial variation. The penalty can also be applied to other INR embeddings, which we evaluate in Sec.~\ref{sec:adaptive}. For classification, we use a one-hot ridge predictor, fitting Eq.~\ref{eq:penalty} for each class with shared $\alpha$ and $\gamma$. We predict the class with the largest linear output.

\section{CoordBench: Measuring Spatial Generalization}
\label{sec:bench}
CoordBench contains $52$ datasets from nine sources~\citep{pdfm2024,cdcplaces2023,sustainbench2021,mosaiks2021,satclip2024,alphaearth2025,worldclim2017,soilgrids2021,pace1997sparse}, with $78$ prediction targets (Table~\ref{tab:bench}). The benchmark suite covers socioeconomic targets such as income and housing ($26$ datasets), environmental quantities such as temperature and soil pH ($10$), and land-cover classes ($16$). Each dataset contains observations at a shared set of coordinates and one or more target columns. We normalize them to a common data format and fold assignments so that new encoders can be compared without rebuilding each source's preprocessing.

\begin{figure}[!t]
	\centering
	\includegraphics[width=0.9\linewidth]{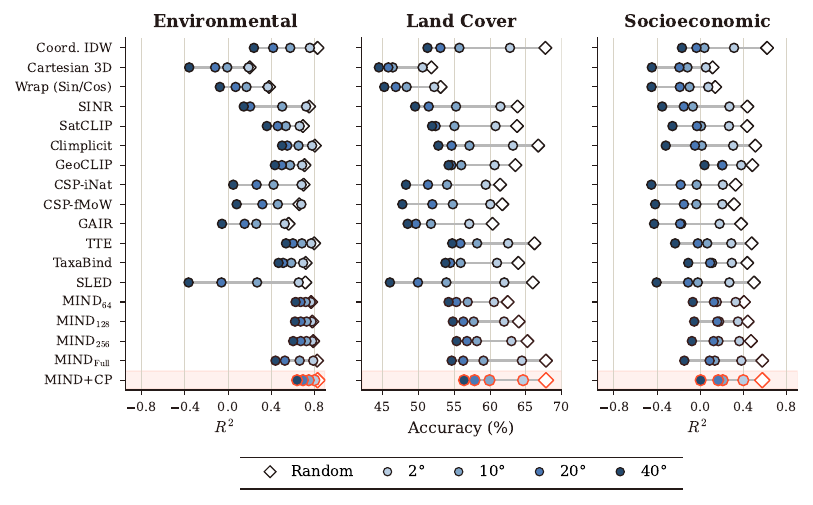}
	\caption{\textbf{CoordBench results by target category.} Each row shows one method's score under random folds (diamond) and regional holdouts with varying block sizes in degrees (darker for larger blocks). \MINDCP has the highest score among INRs at every block size for environmental and land-cover targets. Metrics are the mean scores across five seeds. See Table~\ref{tab:category} for the full results.}
	\label{fig:categories}
\end{figure}

\paragraph{Spatial coverage and resolution.}
The CoordBench datasets differ in label resolution, sampling density, and geographic coverage. WorldClim provides annual temperature and precipitation on a $10$-arc-minute grid~\citep{worldclim2017} and MOSAIKS densely sampled $1$\,km population labels~\citep{mosaiks2021}, while the Africa crop mask~\citep{alphaearth2025} has $2{,}556$ clustered locations. For a coordinate-only encoder, these tasks test both broad patterns in climate and population and local variation in land cover and soil carbon.

\paragraph{Splits.}
In all splits, we assign observations to five random folds. For \emph{random folds}, we allocate points to folds uniformly at random. To evaluate prediction across geographic gaps, CoordBench adds common \emph{regional folds} by spatial blocking, assigning latitude--longitude cells to five folds using a seeded permutation with balanced cell counts. Increasing block cell size from $0.25^\circ$ to $40^\circ$ generally increases train--test distance, but east--west extent shrinks toward the poles, so degrees do not specify a uniform physical holdout size. Appendix Fig.~\ref{fig:spatial-splits} illustrates these folds. Both splits withhold downstream labels from predictor training, while the encoders remain pretrained on globally distributed data. We also perform a buffered holdout experiment and find that performance declines as test points near training points are removed (App.~\ref{app:buffer}). For comparison with existing INR benchmarks, LocBench and TorchSpatial neglect regional holdouts and instead use random $80/20$ splits, even ignoring SustainBench's original country-level splits~\citep{torchspatial2024}.

\paragraph{Evaluation.}
For regression, we floor each dataset's $R^2$ score at $-1$ before averaging across datasets so a few divergent targets do not dominate the aggregate (App.~\ref{app:floor}). All methods use the same datasets at each block size (App.~\ref{app:repro}). We repeat five-fold evaluation with five random seeds to shuffle the fold assignments and report the mean.

\begin{figure}[!t]
	\begin{minipage}[t]{0.50\linewidth}
		\vspace{0pt}
		\captionsetup{type=table}
		\centering\small
\setlength{\tabcolsep}{3pt}
\caption{\textbf{Chunk selection and regularization.}
The \ChunkedPenalty improves regional $R^2$ for \MIND, but not for the non-nested model, GeoCLIP, or permuted \MIND.}
\begin{tabular}{@{}lccc@{}}
\toprule
Method & CP & $10^\circ$ & $40^\circ$ \\
\midrule
\multirow{2}{*}{Non-nested} & $\times$     & \textbf{0.338} & \textbf{0.127} \\
    & $\checkmark$ & 0.330 & 0.114 \\
\midrule
\multirow{2}{*}{GeoCLIP} & $\times$     & \textbf{0.303} & \textbf{0.151} \\
    & $\checkmark$ & 0.303 & 0.148 \\
\midrule
\MIND (chunk sel.) & -- & 0.321 & 0.134 \\
\MIND              & $\checkmark$ & \textbf{0.356} & \textbf{0.184} \\
\midrule
\multirow{2}{*}{\MINDFull (Perm.)} & $\times$     & \textbf{0.289} & \textbf{0.115} \\
    & $\checkmark$ & 0.286 & 0.099 \\
\bottomrule
\end{tabular}
\label{tab:adaptive}

	\end{minipage}\hfill
	\begin{minipage}[t]{0.47\linewidth}
		\vspace{0pt}
		\centering
		\panelLeak
		\caption{\textbf{Held-out distances.} Cumulative distributions of distance to the nearest training point, pooled across CoordBench. Under random folds, $\randwithin\%$ of observations lie within $25$\,km of training data; under $10^\circ$ regional holdout, $\blockbeyond\%$ lie farther than $25$\,km.}
		\label{fig:leak}
	\end{minipage}
\end{figure}

\section{Experiments}
\label{sec:experiments}

We compare \MIND with nine pretrained INR families: SINR~\citep{sinr2023}, SatCLIP~\citep{satclip2024}, Climplicit~\citep{climplicit2025}, GeoCLIP~\citep{geoclip2023}, the two Contrastive Spatial Pretraining (CSP) checkpoints~\citep{csp2023}, trained on iNaturalist~\citep{vanhorn2018inaturalist} and Functional Map of the World (fMoW)~\citep{christie2018fmow}, Geo-aligned Implicit Representations (GAIR)~\citep{gair2025}, Tessellating The Earth (TTE)~\citep{cher2026tte}, TaxaBind~\citep{taxabind2025}, and SLED~\citep{sled2026}. We exclude UniGeoCLIP~\citep{unigeoclip2026} and FAIR-Earth's spherical wavelet models~\citep{cai2025no} because neither releases trained checkpoints or training data. We restrict encoder comparisons to coordinate-only models, excluding the imagery-based AEF and TESSERA v2~\citep{alphaearth2025,feng2026tesserav2scalingpixelwise}. We baseline against the Wrap (Sin/Cos) and Cartesian 3D (XYZ) nonparametric coordinate encoders, as well as Coordinate IDW to measure local interpolation. Coordinate IDW uses distance-weighted k-nearest neighbors, taking a weighted average for regression and a weighted class vote for classification. We fit linear predictors to all learned representations using frozen features and standard ridge regression, except that \MINDCP uses the \ChunkedPenalty described in Eq.~\ref{eq:penalty}. Further details are described in App.~\ref{app:repro}.

\paragraph{Experimental setup.}
We use nested cross-validation to tune the truncation dimension and regularization parameters separately for each target~\citep{cawley2010over}. For each outer test fold, we choose these settings using inner validation folds within the remaining data. Both levels use regional folds for regional holdout and random folds otherwise.

\subsection{Prediction in held-out regions}
\label{sec:what}
\label{sec:results}

Under random folds, $\randwithin\%$ of held-out observations are within $25$\,km of a training point (Fig.~\ref{fig:leak}), and Coordinate IDW outperforms every INR in aggregate regression $R^2$ (Table~\ref{tab:coordbench}). The same effect appears within \MIND: the full embedding performs better than the first $64$ dimensions for nearby test points, but the difference is much smaller beyond $50$\,km. Removing nearby training labels similarly reduces performance more for the full embedding than for the first chunk (App.~\ref{app:buffer}). This suggests that later \MIND dimensions are most useful when nearby labeled data are available, while earlier dimensions are less dependent on local supervision.

Among the pretrained INR baselines, Climplicit performs best under random folds, while GeoCLIP performs best under regional holdouts (Table~\ref{tab:coordbench}). \MINDCP exceeds both, with the highest overall $R^2$ and accuracy among the tested INRs in every split. This result is consistent across fold assignments: \MINDCP outperforms GeoCLIP in $24$ of $25$ split-by-seed comparisons and Climplicit in all $25$ (Table~\ref{tab:coordbench-spread}). It also exceeds Coordinate IDW at every regional block size, while simply concatenating the teachers performs worse than \MINDCP in every evaluated split (App.~\ref{app:teacher-concat}). Overall, \MINDCP remains competitive when nearby training data are available while improving performance as training and test locations become more geographically separated.

\begin{figure}[!t]
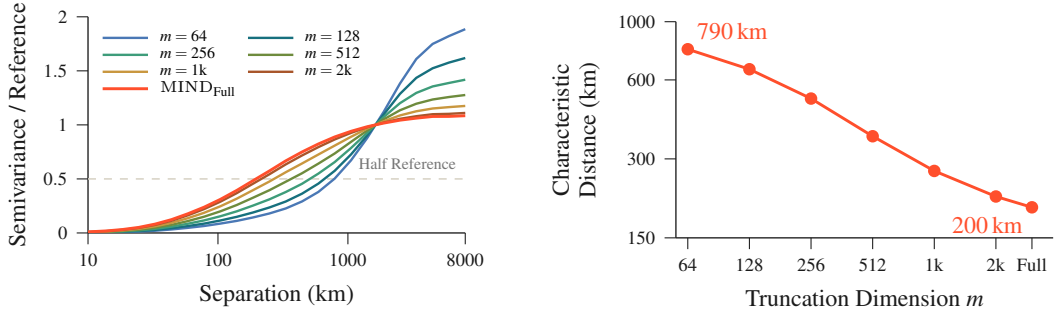

	\centering
	\begin{subfigure}{0.485\linewidth}\panelVariogram\end{subfigure}\hfill
	\begin{subfigure}{0.485\linewidth}\panelHalfSill\end{subfigure}
	\caption{\textbf{Spatial scale across truncations.} Left: semivariance against geographic separation, normalized by the median semivariance over the largest quartile of sampled separations. Curves for more chunks rise at shorter distances. Right: the separation where each curve reaches half this reference value, decreasing from $790$\,km for the first chunk to $200$\,km for the full embedding.}
	\label{fig:scale}
\end{figure}

\subsection{Spatial scale across chunks}
\label{sec:width}
\begin{figure}[!t]
	\centering
	\setlength{\fboxsep}{0pt}\setlength{\fboxrule}{0.5pt}
	\begin{subfigure}{\dimexpr(\linewidth-4pt)/2\relax}\centering\fbox{\includegraphics[width=\dimexpr\linewidth-1pt\relax]{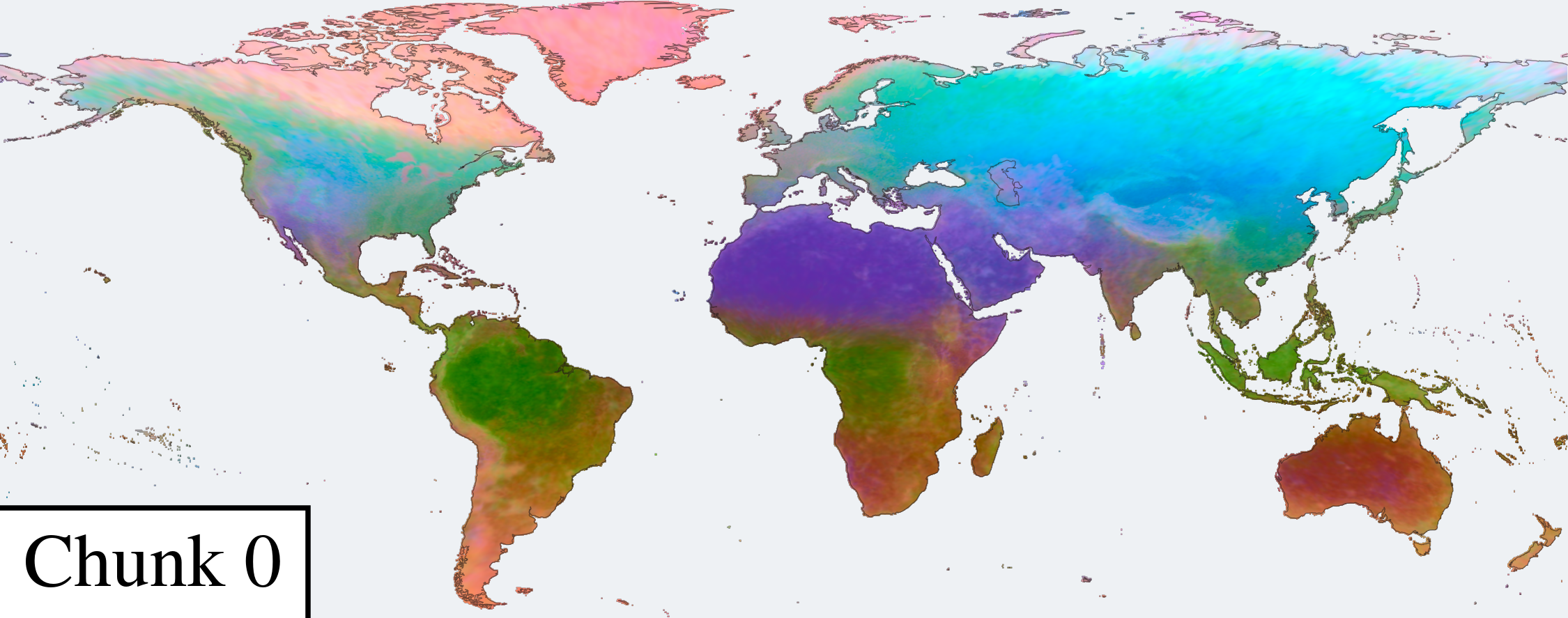}}\end{subfigure}\hspace{4pt}%
	\begin{subfigure}{\dimexpr(\linewidth-4pt)/2\relax}\centering\fbox{\includegraphics[width=\dimexpr\linewidth-1pt\relax]{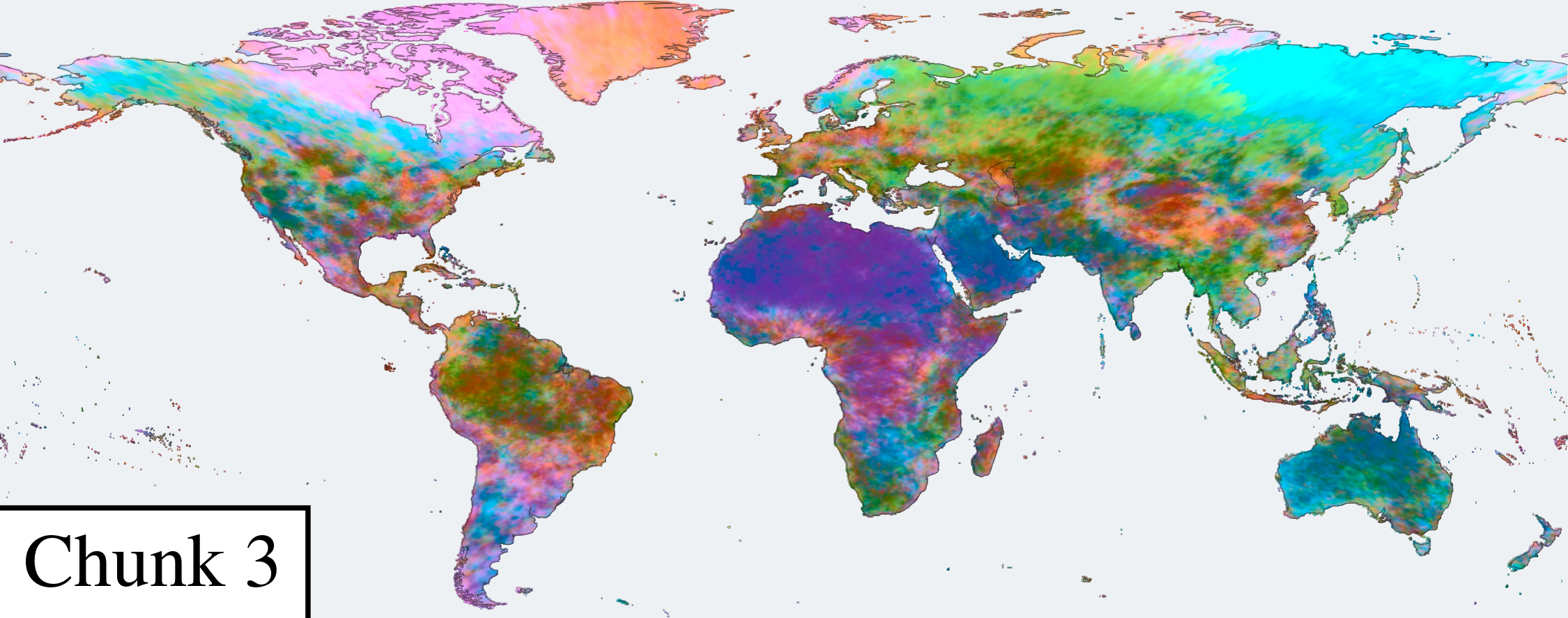}}\end{subfigure}\par\nointerlineskip\vspace{4pt}
	\begin{subfigure}{\dimexpr(\linewidth-4pt)/2\relax}\centering\fbox{\includegraphics[width=\dimexpr\linewidth-1pt\relax]{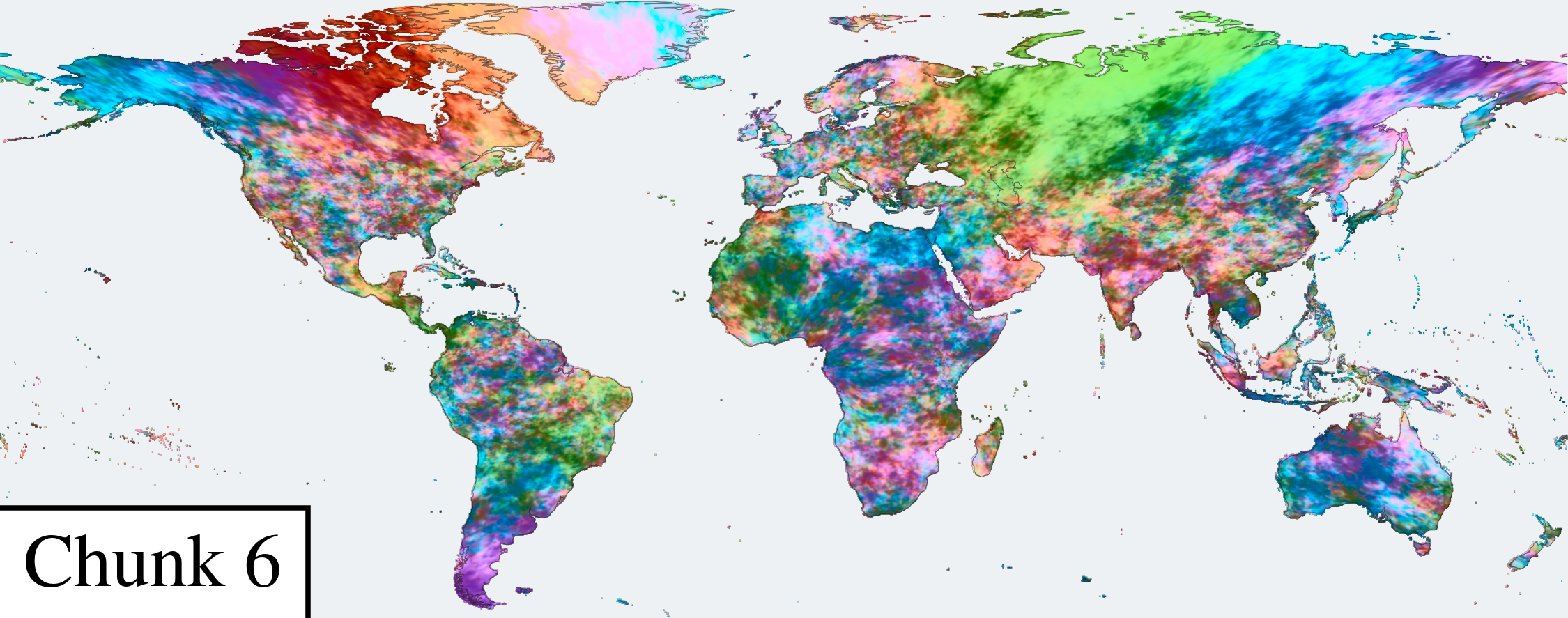}}\end{subfigure}\hspace{4pt}%
	\begin{subfigure}{\dimexpr(\linewidth-4pt)/2\relax}\centering\fbox{\includegraphics[width=\dimexpr\linewidth-1pt\relax]{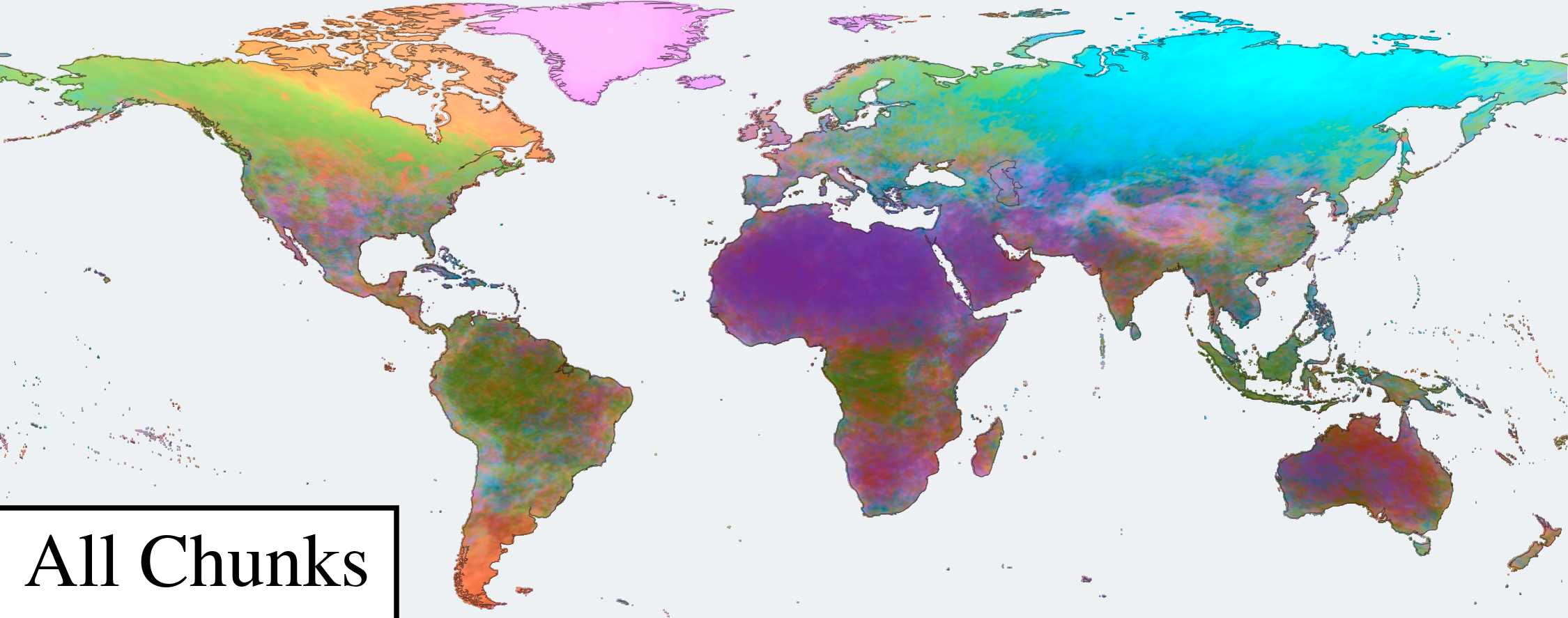}}\end{subfigure}
	\caption{\textbf{Spatial scale across chunks.} Chunk 0 varies gradually across continents (top left), Chunk 3 adds finer regional structure (top right), and Chunk 6 varies at finer scales within regions (bottom left). The full embedding combines variation across spatial scales (bottom right). Each panel maps the top three principal components of the dimensions.}
	\label{fig:bands}
\end{figure}

\MIND's early chunks vary smoothly across broad geographic areas, while later chunks add increasingly local variation (Figs.~\ref{fig:scale} and~\ref{fig:bands}). We hypothesize that the limited capacity of early chunks, together with the spectral bias of neural networks~\citep{fourierfeatures2020,spectralbias2019}, favors broad spatial patterns shared across teachers, such as latitudinal climate gradients, while finer local variation is pushed into later chunks. Notably, nested supervision does not explicitly assign a spatial scale to any chunk but it emerges implicitly through the nested distillation training from specialist teachers.

We quantify this pattern with an empirical \emph{semivariogram}, which measures how quickly embeddings change with geographic distance (Fig.~\ref{fig:scale}). We summarize it using a \emph{characteristic distance}: the separation at which semivariance reaches half its large-distance reference value. Shorter distances indicate finer spatial variation. As more chunks are included, the characteristic distance falls from about $790$\,km for the first $64$ dimensions to $200$\,km for the full embedding. App.~\ref{app:scale} describes the estimator.

Adding chunks improves random-fold regression, while truncating \MIND gives higher $R^2$ for regional holdouts (Table~\ref{tab:coordbench}, Fig.~\ref{fig:dims}). Performance also drops below $64$ dimensions, indicating that shortening the first chunk further loses useful information (App.~\ref{app:scale}). \emph{This coarse-to-fine ordering appears to come from nested supervision rather than representation size alone.} The characteristic distance decreases as chunks are added to \MIND, but remains nearly constant across dimensions in models trained at full width, with or without the addition of feature decorrelation methods like VICReg~\citep{vicreg2022} (App.~\ref{app:scale}). Likewise, separately trained $64$- and $256$-dimensional models underperform \MIND truncated to the same dimensions across all splits (Table~\ref{tab:fixw-main}).

\subsection{Chunk selection, regularization, and spatial ordering}
\label{sec:adaptive}

The \ChunkedPenalty improves regional $R^2$ over per-target chunk selection at $10^\circ$ and $40^\circ$ without reducing \MINDFull's random-fold performance (Table~\ref{tab:adaptive}); at $10^\circ$, chunk selection also performs worse than simply using \MINDdim{128} (Table~\ref{tab:coordbench}). Its benefit depends on \MIND's nested representation. An identical architecture trained only at full dimension performs better than \MINDFull without the penalty ($0.338$ vs. $0.280$ at $10^\circ$), but does not improve with the \ChunkedPenalty, and \MINDCP outperforms this non-nested variant. GeoCLIP likewise does not benefit from the penalty because its dimensions are not nested. Randomly permuting \MIND's dimensions leaves performance unchanged but removes the gain from the \ChunkedPenalty (Table~\ref{tab:adaptive}), suggesting that the improvement comes from the ordering induced by nested supervision rather than regularization alone.

\section{Discussion and Conclusion}
\label{sec:discussion}

\paragraph{Geographic evaluation.}
Coordinate IDW is strongest in regression under random folds, where test points are often close to training observations, but its advantage reverses under regional holdout (Table~\ref{tab:coordbench}). The buffered-exclusion experiment shows that this effect is driven in part by proximity itself (App.~\ref{app:buffer}). Random folds are therefore useful for measuring interpolation near existing observations, but not regional generalization. We recommend reporting IDW, train--test distances, and regional holdouts at scales relevant to deployment. A natural extension is to krige the residuals of an embedding-based predictor~\citep{hengl2007regression}, using nearby labels for local correction and the learned representation farther away.

\paragraph{Spatial scale.}
\MIND adapts the spatial scale of a fixed representation without retraining. Early chunks capture broader patterns, while later chunks add finer variation that can help nearby prediction but transfer less well to unseen regions. Truncation removes later chunks entirely, while the \ChunkedPenalty downweights them, which may explain its higher regional $R^2$ than chunk selection (Table~\ref{tab:adaptive}). The non-nested and permutation experiments suggest that this benefit depends primarily on the learned ordering rather than regularization alone. When spatial validation is impractical, characteristic distance could provide a label-free way to choose an appropriate representation scale.

\paragraph{Limitations and future work.}
CoordBench is geographically imbalanced, with $43$ of $78$ targets from the United States, so evidence is weaker in data-poor regions. Its common format makes it straightforward to add more spatial and temporal datasets. \MIND's learned ordering also depends on its teacher representations and training samples. Broader sampling beyond \MINDSET's current emphasis on cities may improve reconstruction in rural and sparsely sampled areas. \MINDSET's annual AEF embeddings could also be used to study whether a similar ordering emerges over time.

\paragraph{Conclusion.}
Fine-scale features are most useful near labeled data, while smoother features transfer better across larger gaps. \MIND captures both in a single nested representation, allowing downstream models to adjust spatial scale without retraining the encoder. Truncation is useful when the desired scale is known and also reduces storage and probing cost. When it is not, the \ChunkedPenalty provides a softer way to reduce reliance on finer-scale features. CoordBench makes these differences measurable across random and regional holdouts, showing that the best geographic representation depends not only on the target, but also on how far predictions are made from labeled data.

\section*{Reproducibility Statement}
The supplementary material contains training and evaluation scripts, model configuration, and per-dataset results for reproducing the experiments. Appendix~\ref{app:repro} specifies the objective, optimization, predictors, folds, and aggregation. Apps.~\ref{app:bench} and~\ref{app:mindset} describe data provenance. The table and figure scripts derive reported aggregates from these results. The project page is at {\small\href{https://research.taylorgeospatial.org/mind/}{\nolinkurl{research.taylorgeospatial.org/mind}}}. CoordBench is at {\small\href{https://huggingface.co/datasets/taylor-geospatial/CoordBench}{\nolinkurl{huggingface.co/datasets/taylor-geospatial/CoordBench}}}, MIND model checkpoints at {\small\href{https://huggingface.co/taylor-geospatial/MIND}{\nolinkurl{huggingface.co/taylor-geospatial/MIND}}}, the global grid at {\small\href{https://source.coop/tge-labs/mind}{\nolinkurl{source.coop/tge-labs/mind}}}, and the \MINDSET pretraining dataset at {\small\href{https://huggingface.co/datasets/taylor-geospatial/MINDSET}{\nolinkurl{huggingface.co/datasets/taylor-geospatial/MINDSET}}}.

\bibliographystyle{iclr2027_conference}
\begingroup
\urlstyle{rm}
\bibliography{references}
\endgroup

\clearpage
\appendix
\section*{Appendix}

\section{Reproducibility}
\label{app:repro}
\paragraph{Training.}
Training uses AdamW~\citep{loshchilov2019adamw} at a peak learning rate of $3\times10^{-4}$ and weight decay of $0.05$. We use a cosine schedule with $1{,}000$ linear warmup steps, mixed precision, and a batch size of $2048$. The encoder trains for $12{,}000$ steps on one NVIDIA H100 GPU. We train non-nested models at $64$ and $256$ dimensions, two seeds each, with the same teachers, data, step count, and batch size as \MIND.

\paragraph{Architecture.}
The encoder is a residual SIREN~\citep{climplicit2025,siren2020}. It projects latitude and longitude to Equal Earth coordinates, then applies a sinusoidal input layer of width $\FullDim$ with input-frequency multiplier $30$ and twelve residual sinusoidal blocks. The Equal Earth projection is discontinuous at the antimeridian, where longitude wraps between east and west.

\paragraph{Teachers and objective.}
The frozen teachers are AEF ($64$ dimensions, averaged over its release years), Climplicit ($1024$), GeoCLIP ($512$), and SINR ($256$). AEF targets are precomputed, and the last three teachers are evaluated from coordinates during distillation. We center each teacher's channels and scale them by one root-mean-square value following PHI Standardization (PHI-S)~\citep{phis2024}. The reconstruction loss is defined in Eq.~\ref{eq:reconstruction}. The normalized mean squared error term equals $2(1-\bar{\mathbf{u}}^\top\bar{\mathbf{v}})/d_t$, so the combined loss rescales cosine distance according to teacher dimension. Linear heads reconstruct the targets at dimensions $\mathcal{M}=\{64,128,256,512,1024,2048,d\}$, with full dimension $d=\FullDim$. Averaging the losses over $\mathcal{M}\setminus\{d\}$ keeps the total weighting of the smaller representations independent of their number~\citep{smec2025}. The heads are discarded after training, and evaluation uses the corresponding leading chunks of the encoder output. The teacher pool was selected using random-fold scores before the regional-holdout experiments.

\paragraph{Penalty selection.}
The \ChunkedPenalty uses chunk boundaries at dimensions $64,128,256,512,1024,$ and $2048$ and increases the coefficient penalty by a factor $\gamma$ for each successive chunk. We search $\gamma\in\{3,10,100,1000\}$, with standard ridge regression also included among the candidates. Both truncation and the penalty tune two parameters, $(m,\alpha)$ and $(\alpha,\gamma)$, respectively, using nested cross-validation. For an embedding of dimension $d$, only the boundaries below $d$ apply, so GeoCLIP ($d=512$) uses $\{64,128,256\}$ and no reordering of its dimensions is performed.

\paragraph{Released embeddings.}
We release land embeddings on a global $0.01^\circ$ grid, about $1$\,km at the equator, with $8$-bit quantization. Compared with direct model evaluation, decoded embeddings have cosine similarity $0.99997$ and root-mean-square error $0.022$ per dimension.

\paragraph{Teacher concatenation.}
\label{app:teacher-concat}
We concatenate the frozen Climplicit, GeoCLIP, and SINR embeddings and fit the same ridge predictor. AEF is excluded because querying its embeddings requires access to the raster. At $40^\circ$, the concatenation has $R^2=-0.076$, compared with $0.184$ for \MINDCP (Table~\ref{tab:coordbench-spread}).

\paragraph{Splits.}
We assign each point to a block of size $c$, permute the unique blocks with a fixed seed, and assign successive blocks cyclically to five folds. Unless stated otherwise, results average five fold assignments generated with different random seeds. Blocks have no exclusion zones around their boundaries, so larger blocks generally increase train--test distance but also change which regions are held out. Buffered exclusion removes training data within a radius of fixed held-out sites~\citep{shen2024bloo,shen2026blisco}. App.~\ref{app:buffer} reports this comparison. All sources, including DHS and PDFM, use the common random and regional cross-validation recipe rather than their supplied splits.

\paragraph{Datasets by split.}
We exclude datasets whose median distance from test points to their nearest training points is unchanged from the next smaller block size in the sweep, which also includes $5^\circ$. We also exclude datasets without spatial inner-validation results. Ethiopia Crops is excluded at $10^\circ$, $20^\circ$, and $40^\circ$; California Housing at $20^\circ$ and $40^\circ$. Otherwise, all methods use the same datasets for each split recipe.

\paragraph{Downstream predictors.}
Before fitting each predictor, we center and scale features dimension-wise using means and standard deviations from the corresponding training split, applying the same transformation to validation and test features. We solve ridge regression in closed form using float64 over penalties from $10^{-4}$ to $10^{6}$, spaced at half-unit intervals on a base-ten logarithmic scale. Table~\ref{tab:category} reports results by target category.

For Coordinate IDW, we sweep $k \in \{1, 2, 5, 10, 20, 35, 50, 100, 200, 500\}$ with the folds, datasets, and aggregation held fixed. We fix $k=50$ for all reported comparisons. This choice is within $0.04$ in $R^2$ and $0.5$ accuracy points of the best neighbor count in every split. We weight neighbors by inverse great-circle distance with power $p=1$.

\begin{figure}[t]
	\centering
	\includegraphics[width=\linewidth]{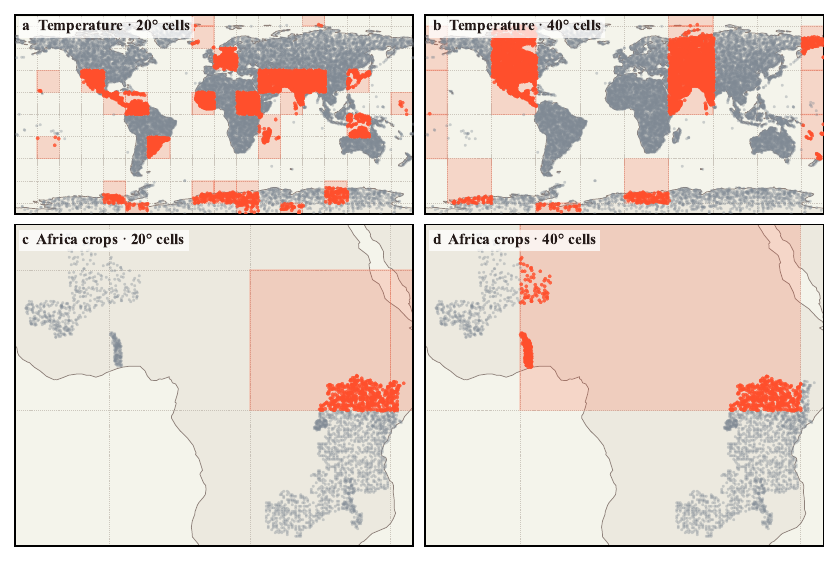}
	\caption{\textbf{Regional folds at $20^\circ$ and $40^\circ$.} WorldClim temperature has $18{,}068$ globally distributed observations (top). Africa crop labels are sparse and clustered, with $2{,}556$ observations (bottom). Coral points and shaded cells show one held-out fold, and gray points are training observations. Both use fold $0$ of the five-fold assignment with seed $0$. Cells are assigned to folds independently at each block size, so the held-out regions change.}
	\label{fig:spatial-splits}
\end{figure}

\section{CoordBench provenance}
\label{app:bench}
We convert CoordBench datasets from their raw sources to tabular format with \textit{latitude}, \textit{longitude}, \textit{timestamp}, and \textit{split} columns. Not all datasets include time information or official splits. The six Demographic and Health Surveys datasets include deliberate displacement of survey-cluster coordinates, up to $2$\,km in urban areas and $5$\,km in rural areas~\citep{burgert2013dhs}. This limits the interpretation of sub-$5$\,km distance statistics on those datasets. The dataset card records provenance, citation, and license status for each table. Licenses differ across sources.

\section{\MINDSET, the pretraining dataset}
\label{app:mindset}
\MIND is trained on \MINDSET, a tabular dataset of teacher embeddings at $12{,}099{,}072$ land coordinates, released alongside the model. The coordinates are AEF pixels sampled densely in and around cities. Blocks about $2.5$\,km wide are placed inside the bounding boxes of GeoNames cities15000~\citep{geonames_database} (roughly $34$k cities with a population of at least $15{,}000$), and cities are drawn with probability proportional to the square root of population. Each block contributes $768$ pixels at a mean spacing near $90$\,m, so training batches contain many close pairs. A pixel is kept only if it is valid in all nine annual AEF mosaics ($2017$--$2025$). The released latitude/longitude-only model distills the mean of a point's nine annual AEF embeddings. The dataset card records the layout, provenance, and license of each upstream model.

\section{Characteristic scale of the embedding field}
\label{app:scale}
We compute the empirical semivariogram from $20$k reference land locations and coordinate pairs at $48$ distances measured along Earth's surface. The curve is still increasing at $8000$\,km. We use the median semivariance over the largest quartile of measured separations as a reference value and define the characteristic distance as the separation where the curve reaches half this value. We interpolate this distance on logarithmic axes. All dimensions and objectives use the same locations, separations, and estimator. Broad geographic patterns also appear in later chunks, so spatial scales overlap across chunks.

\paragraph{Effect of nested supervision.}
To isolate the nesting objective, we hold the encoder architecture fixed and vary only the objective. The characteristic distances of three nested variants decrease by factors of $2.7\times$ to $4.0\times$ from $64$ dimensions to the full embedding. The two variants trained only on the full embedding stay near $230$\,km at every truncation, and the two VICReg~\citep{vicreg2022} variants stay near $117$\,km. The VICReg controls use a covariance penalty alone or together with a variance penalty. At $40^\circ$, the nested variant has positive scores over a range of dimensions under all five fold-assignment seeds and the highest mean score of any variant, while the VICReg controls stay negative.

\paragraph{Truncation below 64 dimensions.}
$R^2$ is lower at $16$ and $32$ dimensions than at $64$ under every split. At $10^\circ$, $R^2$ increases from $0.128$ at $16$ dimensions to $0.309$ at $64$ dimensions.

\begin{table}[t]
	\centering
	\captionsetup{justification=raggedright,singlelinecheck=false}
	\begin{minipage}[t]{0.48\linewidth}
		\vspace{0pt}
			\centering\fontsize{8}{9.6}\selectfont
	\setlength{\tabcolsep}{2.5pt}
	\caption{\textbf{Removing nearby training observations.} Buffered exclusion lowers $R^2$ more than removing the same number of training observations at random. \MINDCP exceeds GeoCLIP at every radius.}
	\label{tab:buffer}
	\begin{tabular}{@{}ccccc@{}}
		\toprule
		& & \multicolumn{2}{c}{\MINDCP} & GeoCLIP \\
		\cmidrule(lr){3-4}
        \cmidrule(lr){5-5}
		\shortstack{Radius\\(km)} & \shortstack{Median\\(km)} & Buffered & Random & Buffered \\
		\midrule
		$0$   & $8$   & 0.68 & 0.68 & 0.56 \\
		$10$  & $15$  & 0.61 & 0.67 & 0.52 \\
		$25$  & $42$  & 0.49 & 0.63 & 0.42 \\
		$50$  & $194$ & 0.29 & 0.57 & 0.23 \\
		$100$ & $632$ & $0.10$ & 0.47 & $0.04$ \\
		\bottomrule
	\end{tabular}

	\end{minipage}\hfill
	\begin{minipage}[t]{0.48\linewidth}
		\vspace{0pt}
		\centering\fontsize{8}{9.6}\selectfont
	\setlength{\tabcolsep}{3pt}
	\caption{\textbf{Share of regression scores at the $R^2$ floor (\%).} Percentage of scores at or below $-1$ across targets and fold-assignment seeds, using the datasets in Table~\ref{tab:coordbench}.\label{tab:floor}}
	\begin{tabular}{@{}lccccc@{}}
		\toprule
		Method & Random & $2^\circ$ & $10^\circ$ & $20^\circ$ & $40^\circ$ \\
		\midrule
		Coord. IDW & 0.0 & 0.0 & 0.0 & 0.0 & 0.0 \\
		\midrule
		SINR & 0.0 & 0.0 & 3.9 & 5.9 & 10.5 \\
		SLED & 0.0 & 0.0 & 1.6 & 3.3 & 8.5 \\
		CSP-fMoW & 0.3 & 0.6 & 3.2 & 3.9 & 10.5 \\
		Climplicit & 0.0 & 0.0 & 1.6 & 0.0 & 5.9 \\
		GeoCLIP & 0.0 & 0.0 & 1.6 & 0.0 & 1.0 \\
		\midrule
		\MINDdim{64} & 0.0 & 0.3 & 1.6 & 0.0 & 2.6 \\
		\MINDdim{128} & 0.0 & 0.3 & 1.6 & 0.0 & 2.0 \\
		\MINDFull & 0.0 & 0.0 & 0.0 & 0.0 & 3.0 \\
		\MINDCP & 0.0 & 0.0 & 1.6 & 0.0 & 1.3 \\
		\bottomrule
	\end{tabular}

	\end{minipage}
\end{table}

\section{Buffered exclusion with fixed test points}
\label{app:buffer}

Regional block size changes both train--test distance and which regions are held out. To isolate the effect of distance, we instead fix the test points from random-split fold and remove training observations within a great-circle radius of each test point. We evaluate radii from $10$ to $400$\,km and compare against a control that removes the same number of training observations uniformly at random. Because exclusion regions can overlap, the nearest remaining training point may be farther than the exclusion radius. Table~\ref{tab:buffer} reports the actual median train--test distance.

Removing nearby observations hurts performance more than removing the same number at random. At a $50$\,km exclusion radius, \MINDCP drops to $R^2=0.294$, compared with $0.572$ under random deletion, while still outperforming GeoCLIP at every radius (Table~\ref{tab:buffer}). The same pattern appears within \MIND, where removing observations within $25$\,km reduces \MINDFull from $0.676$ to $0.489$ compared with $0.534$ to $0.468$ for \MINDdim{64}. Coordinate IDW is even more sensitive, dropping from $0.698$ to $0.106$ at $100$\,km. This shows that nearby training data matter even when distance is measured directly in kilometers rather than by latitude--longitude regions.

\section{Incidence of the $R^2$ floor}
\label{app:floor}
We replace target-level $R^2$ values below $-1$ with $-1$ before averaging. This limits the contribution of large negative scores but obscures differences below the threshold. Table~\ref{tab:floor} reports the percentage of regression scores at or below $-1$ across targets and fold-assignment seeds, using the datasets in Table~\ref{tab:coordbench}. Across split settings, this percentage is at most $1.6\%$ for \MINDCP, $2.6\%$ for \MINDdim{64}, and $1.6\%$ for GeoCLIP. For SINR, it reaches $10.5\%$ at $40^\circ$.

\begin{table}[p]\centering\scriptsize
	\setlength{\tabcolsep}{1.6pt}
    \caption{\textbf{CoordBench results by target category.} Values from Fig.~\ref{fig:categories} are $R^2$ for environmental and socioeconomic targets and accuracy (\%) for land cover. Header counts denote datasets.
\label{tab:category}}
	\begin{tabular}{@{}lccccccccccccccc@{}}
		\toprule
		& \multicolumn{5}{c}{Environmental ($10$)} & \multicolumn{5}{c}{Land Cover ($16$)} & \multicolumn{5}{c}{Socioeconomic ($26$)} \\
		\cmidrule(lr){2-6}\cmidrule(lr){7-11}\cmidrule(l){12-16}
		Method & Rand. & $2^\circ$ & $10^\circ$ & $20^\circ$ & $40^\circ$ & Rand. & $2^\circ$ & $10^\circ$ & $20^\circ$ & $40^\circ$ & Rand. & $2^\circ$ & $10^\circ$ & $20^\circ$ & $40^\circ$ \\
		\midrule
		Coord. IDW & \textbf{0.827} & 0.756 & 0.575 & 0.416 & 0.238 & 67.7 & 62.8 & 55.7 & 53.0 & 51.2 & \textbf{0.616} & 0.311 & 0.037 & -0.036 & -0.170 \\
		\midrule
		Cartesian 3D & 0.200 & 0.189 & -0.008 & -0.120 & -0.360 & 51.8 & 50.6 & 46.4 & 45.8 & 44.4 & 0.113 & 0.050 & -0.119 & -0.194 & -0.449 \\
		Wrap (Sin/Cos) & 0.379 & 0.368 & 0.170 & 0.069 & -0.076 & 53.1 & 52.2 & 48.3 & 46.8 & 45.2 & 0.137 & 0.073 & -0.100 & -0.190 & -0.452 \\
		\midrule
		SINR & 0.750 & 0.721 & 0.500 & 0.204 & 0.145 & 63.8 & 61.4 & 55.2 & 51.4 & 49.5 & 0.434 & 0.269 & -0.069 & -0.153 & -0.353 \\
		SatCLIP & 0.691 & 0.661 & 0.534 & 0.458 & 0.358 & 63.8 & 60.8 & 55.0 & 52.4 & 51.9 & 0.432 & 0.265 & 0.006 & -0.032 & -0.258 \\
		Climplicit & 0.802 & 0.776 & 0.651 & 0.547 & 0.499 & 66.7 & 63.2 & 57.1 & 54.6 & 52.8 & 0.508 & 0.303 & 0.013 & -0.050 & -0.320 \\
		GeoCLIP & 0.706 & 0.684 & 0.572 & 0.496 & 0.433 & 63.5 & 60.6 & 56.0 & 54.6 & 54.2 & 0.480 & 0.378 & \textit{0.200} & \textbf{0.203} & \textbf{0.038} \\
		CSP-iNat & 0.698 & 0.680 & 0.420 & 0.263 & 0.047 & 61.4 & 59.4 & 54.0 & 51.3 & 48.2 & 0.325 & 0.208 & -0.034 & -0.183 & -0.453 \\
		CSP-fMoW & 0.658 & 0.677 & 0.460 & 0.317 & 0.079 & 61.7 & 60.0 & 54.8 & 51.9 & 47.7 & 0.310 & 0.203 & -0.037 & -0.153 & -0.418 \\
		GAIR & 0.559 & 0.523 & 0.260 & 0.152 & -0.057 & 60.3 & 57.1 & 51.7 & 49.6 & 48.5 & 0.377 & 0.178 & -0.176 & -0.187 & -0.428 \\
		TTE & 0.796 & 0.769 & 0.682 & 0.599 & 0.535 & 66.2 & 62.5 & 58.2 & 55.8 & 54.7 & 0.474 & 0.286 & 0.063 & -0.024 & -0.233 \\
		TaxaBind & 0.717 & 0.694 & 0.584 & 0.508 & 0.466 & 63.9 & 61.0 & 55.9 & 54.4 & 53.7 & 0.433 & 0.291 & 0.108 & 0.090 & -0.112 \\
		SLED & 0.714 & 0.653 & 0.268 & -0.062 & -0.366 & 66.0 & 62.0 & 53.9 & 49.9 & 46.0 & 0.496 & 0.268 & -0.024 & -0.113 & -0.405 \\
		\midrule
		\MINDdim{64} & 0.768 & 0.763 & 0.717 & 0.670 & \textit{0.624} & 62.5 & 60.5 & 56.9 & 55.3 & 54.2 & 0.402 & 0.326 & 0.152 & 0.125 & -0.070 \\
		\MINDdim{128} & 0.778 & 0.772 & \textit{0.726} & 0.671 & 0.618 & 64.0 & 62.0 & 57.7 & 56.3 & 54.8 & 0.438 & 0.347 & 0.173 & 0.156 & -0.057 \\
		\MINDdim{256} & 0.787 & 0.777 & 0.724 & \textit{0.672} & 0.601 & 65.2 & 63.0 & 58.1 & \textit{56.8} & \textit{55.3} & 0.468 & 0.360 & 0.167 & 0.122 & -0.078 \\
		\MINDFull & 0.823 & \textit{0.787} & 0.661 & 0.525 & 0.438 & \textit{67.8} & \textit{64.4} & \textit{59.1} & 56.2 & 54.6 & 0.572 & \textit{0.379} & 0.133 & 0.086 & -0.148 \\
		\MINDCP & \textit{0.827} & \textbf{0.798} & \textbf{0.746} & \textbf{0.692} & \textbf{0.638} & \textbf{67.8} & \textbf{64.6} & \textbf{59.9} & \textbf{57.8} & \textbf{56.3} & \textit{0.572} & \textbf{0.397} & \textbf{0.207} & \textit{0.165} & \textit{0.003} \\
		\bottomrule
	\end{tabular}

\end{table}

\begin{table*}[p]\centering\scriptsize
	\setlength{\tabcolsep}{2pt}
	\caption{\textbf{CoordBench results with mean $\pm$ std across the 5 random seeds.}}
	\label{tab:coordbench-spread}
	\begin{tabular}{@{}lcccccccccc@{}}
		\toprule
		& \multicolumn{5}{c}{$R^2$} & \multicolumn{5}{c}{Accuracy (\%)} \\
		\cmidrule(lr){2-6}\cmidrule(l){7-11}
		Method & Random & $2^\circ$ & $10^\circ$ & $20^\circ$ & $40^\circ$ & Random & $2^\circ$ & $10^\circ$ & $20^\circ$ & $40^\circ$ \\
		\midrule
		Coord. IDW & \shortstack[c]{\textbf{0.675}\\[-1pt]$\pm$ 0.001} & \shortstack[c]{0.435\\[-1pt]$\pm$ 0.007} & \shortstack[c]{0.187\\[-1pt]$\pm$ 0.018} & \shortstack[c]{0.093\\[-1pt]$\pm$ 0.026} & \shortstack[c]{-0.053\\[-1pt]$\pm$ 0.027} & \shortstack[c]{\textit{67.7}\\[-1pt]$\pm$ 0.1} & \shortstack[c]{62.8\\[-1pt]$\pm$ 0.3} & \shortstack[c]{55.7\\[-1pt]$\pm$ 0.4} & \shortstack[c]{53.0\\[-1pt]$\pm$ 0.5} & \shortstack[c]{51.2\\[-1pt]$\pm$ 0.5} \\
		\midrule
		Cartesian 3D & \shortstack[c]{0.137\\[-1pt]$\pm$ 0.000} & \shortstack[c]{0.089\\[-1pt]$\pm$ 0.012} & \shortstack[c]{-0.088\\[-1pt]$\pm$ 0.015} & \shortstack[c]{-0.173\\[-1pt]$\pm$ 0.041} & \shortstack[c]{-0.423\\[-1pt]$\pm$ 0.045} & \shortstack[c]{51.8\\[-1pt]$\pm$ 0.0} & \shortstack[c]{50.6\\[-1pt]$\pm$ 0.4} & \shortstack[c]{46.4\\[-1pt]$\pm$ 0.4} & \shortstack[c]{45.8\\[-1pt]$\pm$ 0.4} & \shortstack[c]{44.4\\[-1pt]$\pm$ 0.9} \\
		Wrap (Sin/Cos) & \shortstack[c]{0.204\\[-1pt]$\pm$ 0.000} & \shortstack[c]{0.155\\[-1pt]$\pm$ 0.012} & \shortstack[c]{-0.025\\[-1pt]$\pm$ 0.014} & \shortstack[c]{-0.116\\[-1pt]$\pm$ 0.044} & \shortstack[c]{-0.344\\[-1pt]$\pm$ 0.047} & \shortstack[c]{53.1\\[-1pt]$\pm$ 0.0} & \shortstack[c]{52.2\\[-1pt]$\pm$ 0.3} & \shortstack[c]{48.3\\[-1pt]$\pm$ 0.1} & \shortstack[c]{46.8\\[-1pt]$\pm$ 0.6} & \shortstack[c]{45.2\\[-1pt]$\pm$ 1.0} \\
		\midrule
		SINR & \shortstack[c]{0.522\\[-1pt]$\pm$ 0.000} & \shortstack[c]{0.394\\[-1pt]$\pm$ 0.010} & \shortstack[c]{0.089\\[-1pt]$\pm$ 0.051} & \shortstack[c]{-0.051\\[-1pt]$\pm$ 0.049} & \shortstack[c]{-0.211\\[-1pt]$\pm$ 0.058} & \shortstack[c]{63.8\\[-1pt]$\pm$ 0.1} & \shortstack[c]{61.4\\[-1pt]$\pm$ 0.3} & \shortstack[c]{55.2\\[-1pt]$\pm$ 0.6} & \shortstack[c]{51.4\\[-1pt]$\pm$ 0.7} & \shortstack[c]{49.5\\[-1pt]$\pm$ 0.7} \\
		SatCLIP & \shortstack[c]{0.504\\[-1pt]$\pm$ 0.000} & \shortstack[c]{0.375\\[-1pt]$\pm$ 0.012} & \shortstack[c]{0.153\\[-1pt]$\pm$ 0.016} & \shortstack[c]{0.108\\[-1pt]$\pm$ 0.022} & \shortstack[c]{-0.082\\[-1pt]$\pm$ 0.102} & \shortstack[c]{63.8\\[-1pt]$\pm$ 0.0} & \shortstack[c]{60.8\\[-1pt]$\pm$ 0.3} & \shortstack[c]{55.0\\[-1pt]$\pm$ 0.4} & \shortstack[c]{52.4\\[-1pt]$\pm$ 0.4} & \shortstack[c]{51.9\\[-1pt]$\pm$ 0.6} \\
		Climplicit & \shortstack[c]{0.590\\[-1pt]$\pm$ 0.000} & \shortstack[c]{0.435\\[-1pt]$\pm$ 0.010} & \shortstack[c]{0.190\\[-1pt]$\pm$ 0.012} & \shortstack[c]{0.120\\[-1pt]$\pm$ 0.023} & \shortstack[c]{-0.086\\[-1pt]$\pm$ 0.097} & \shortstack[c]{66.7\\[-1pt]$\pm$ 0.1} & \shortstack[c]{63.2\\[-1pt]$\pm$ 0.4} & \shortstack[c]{57.1\\[-1pt]$\pm$ 0.5} & \shortstack[c]{54.6\\[-1pt]$\pm$ 0.5} & \shortstack[c]{52.8\\[-1pt]$\pm$ 0.5} \\
		GeoCLIP & \shortstack[c]{0.543\\[-1pt]$\pm$ 0.000} & \shortstack[c]{0.463\\[-1pt]$\pm$ 0.012} & \shortstack[c]{0.303\\[-1pt]$\pm$ 0.012} & \shortstack[c]{0.287\\[-1pt]$\pm$ 0.011} & \shortstack[c]{\textit{0.151}\\[-1pt]$\pm$ 0.091} & \shortstack[c]{63.5\\[-1pt]$\pm$ 0.0} & \shortstack[c]{60.6\\[-1pt]$\pm$ 0.2} & \shortstack[c]{56.0\\[-1pt]$\pm$ 0.3} & \shortstack[c]{54.6\\[-1pt]$\pm$ 0.5} & \shortstack[c]{54.2\\[-1pt]$\pm$ 0.6} \\
		CSP-iNat & \shortstack[c]{0.429\\[-1pt]$\pm$ 0.000} & \shortstack[c]{0.339\\[-1pt]$\pm$ 0.011} & \shortstack[c]{0.092\\[-1pt]$\pm$ 0.035} & \shortstack[c]{-0.055\\[-1pt]$\pm$ 0.027} & \shortstack[c]{-0.310\\[-1pt]$\pm$ 0.050} & \shortstack[c]{61.4\\[-1pt]$\pm$ 0.0} & \shortstack[c]{59.4\\[-1pt]$\pm$ 0.4} & \shortstack[c]{54.0\\[-1pt]$\pm$ 0.4} & \shortstack[c]{51.3\\[-1pt]$\pm$ 0.6} & \shortstack[c]{48.2\\[-1pt]$\pm$ 1.0} \\
		CSP-fMoW & \shortstack[c]{0.407\\[-1pt]$\pm$ 0.019} & \shortstack[c]{0.335\\[-1pt]$\pm$ 0.004} & \shortstack[c]{0.101\\[-1pt]$\pm$ 0.015} & \shortstack[c]{-0.018\\[-1pt]$\pm$ 0.023} & \shortstack[c]{-0.276\\[-1pt]$\pm$ 0.038} & \shortstack[c]{61.7\\[-1pt]$\pm$ 0.0} & \shortstack[c]{60.0\\[-1pt]$\pm$ 0.3} & \shortstack[c]{54.8\\[-1pt]$\pm$ 0.4} & \shortstack[c]{51.9\\[-1pt]$\pm$ 0.8} & \shortstack[c]{47.7\\[-1pt]$\pm$ 0.7} \\
		GAIR & \shortstack[c]{0.427\\[-1pt]$\pm$ 0.000} & \shortstack[c]{0.274\\[-1pt]$\pm$ 0.017} & \shortstack[c]{-0.055\\[-1pt]$\pm$ 0.038} & \shortstack[c]{-0.090\\[-1pt]$\pm$ 0.021} & \shortstack[c]{-0.322\\[-1pt]$\pm$ 0.051} & \shortstack[c]{60.3\\[-1pt]$\pm$ 0.0} & \shortstack[c]{57.1\\[-1pt]$\pm$ 0.4} & \shortstack[c]{51.7\\[-1pt]$\pm$ 0.2} & \shortstack[c]{49.6\\[-1pt]$\pm$ 0.4} & \shortstack[c]{48.5\\[-1pt]$\pm$ 1.1} \\
		TTE & \shortstack[c]{0.563\\[-1pt]$\pm$ 0.000} & \shortstack[c]{0.420\\[-1pt]$\pm$ 0.013} & \shortstack[c]{0.235\\[-1pt]$\pm$ 0.013} & \shortstack[c]{0.154\\[-1pt]$\pm$ 0.017} & \shortstack[c]{-0.014\\[-1pt]$\pm$ 0.072} & \shortstack[c]{66.2\\[-1pt]$\pm$ 0.1} & \shortstack[c]{62.5\\[-1pt]$\pm$ 0.4} & \shortstack[c]{58.2\\[-1pt]$\pm$ 0.6} & \shortstack[c]{55.8\\[-1pt]$\pm$ 0.7} & \shortstack[c]{54.7\\[-1pt]$\pm$ 0.6} \\
		TaxaBind & \shortstack[c]{0.512\\[-1pt]$\pm$ 0.000} & \shortstack[c]{0.402\\[-1pt]$\pm$ 0.014} & \shortstack[c]{0.240\\[-1pt]$\pm$ 0.008} & \shortstack[c]{0.210\\[-1pt]$\pm$ 0.009} & \shortstack[c]{0.053\\[-1pt]$\pm$ 0.090} & \shortstack[c]{63.9\\[-1pt]$\pm$ 0.0} & \shortstack[c]{61.0\\[-1pt]$\pm$ 0.3} & \shortstack[c]{55.9\\[-1pt]$\pm$ 0.1} & \shortstack[c]{54.4\\[-1pt]$\pm$ 0.5} & \shortstack[c]{53.7\\[-1pt]$\pm$ 0.4} \\
		SLED & \shortstack[c]{0.557\\[-1pt]$\pm$ 0.001} & \shortstack[c]{0.375\\[-1pt]$\pm$ 0.013} & \shortstack[c]{0.057\\[-1pt]$\pm$ 0.013} & \shortstack[c]{-0.099\\[-1pt]$\pm$ 0.023} & \shortstack[c]{-0.394\\[-1pt]$\pm$ 0.101} & \shortstack[c]{66.0\\[-1pt]$\pm$ 0.1} & \shortstack[c]{62.0\\[-1pt]$\pm$ 0.5} & \shortstack[c]{53.9\\[-1pt]$\pm$ 0.4} & \shortstack[c]{49.9\\[-1pt]$\pm$ 0.7} & \shortstack[c]{46.0\\[-1pt]$\pm$ 1.0} \\
		Teacher Concat. & \shortstack[c]{0.634\\[-1pt]$\pm$ 0.000} & \shortstack[c]{\textit{0.496}\\[-1pt]$\pm$ 0.036} & \shortstack[c]{0.259\\[-1pt]$\pm$ 0.019} & \shortstack[c]{0.156\\[-1pt]$\pm$ 0.045} & \shortstack[c]{-0.076\\[-1pt]$\pm$ 0.100} & \shortstack[c]{67.4\\[-1pt]$\pm$ 0.1} & \shortstack[c]{64.0\\[-1pt]$\pm$ 0.5} & \shortstack[c]{58.6\\[-1pt]$\pm$ 0.7} & \shortstack[c]{55.7\\[-1pt]$\pm$ 0.8} & \shortstack[c]{54.2\\[-1pt]$\pm$ 0.3} \\
		\midrule
		\MINDdim{64} & \shortstack[c]{0.504\\[-1pt]$\pm$ 0.000} & \shortstack[c]{0.447\\[-1pt]$\pm$ 0.014} & \shortstack[c]{0.309\\[-1pt]$\pm$ 0.011} & \shortstack[c]{0.281\\[-1pt]$\pm$ 0.022} & \shortstack[c]{0.128\\[-1pt]$\pm$ 0.095} & \shortstack[c]{62.5\\[-1pt]$\pm$ 0.0} & \shortstack[c]{60.5\\[-1pt]$\pm$ 0.4} & \shortstack[c]{56.9\\[-1pt]$\pm$ 0.5} & \shortstack[c]{55.3\\[-1pt]$\pm$ 0.6} & \shortstack[c]{54.2\\[-1pt]$\pm$ 0.6} \\
		\MINDdim{128} & \shortstack[c]{0.533\\[-1pt]$\pm$ 0.000} & \shortstack[c]{0.465\\[-1pt]$\pm$ 0.014} & \shortstack[c]{\textit{0.327}\\[-1pt]$\pm$ 0.010} & \shortstack[c]{\textit{0.303}\\[-1pt]$\pm$ 0.014} & \shortstack[c]{0.136\\[-1pt]$\pm$ 0.108} & \shortstack[c]{64.0\\[-1pt]$\pm$ 0.1} & \shortstack[c]{62.0\\[-1pt]$\pm$ 0.3} & \shortstack[c]{57.7\\[-1pt]$\pm$ 0.3} & \shortstack[c]{56.3\\[-1pt]$\pm$ 0.4} & \shortstack[c]{54.8\\[-1pt]$\pm$ 0.6} \\
		\MINDdim{256} & \shortstack[c]{0.556\\[-1pt]$\pm$ 0.000} & \shortstack[c]{0.476\\[-1pt]$\pm$ 0.014} & \shortstack[c]{0.322\\[-1pt]$\pm$ 0.009} & \shortstack[c]{0.279\\[-1pt]$\pm$ 0.013} & \shortstack[c]{0.116\\[-1pt]$\pm$ 0.097} & \shortstack[c]{65.2\\[-1pt]$\pm$ 0.1} & \shortstack[c]{63.0\\[-1pt]$\pm$ 0.2} & \shortstack[c]{58.1\\[-1pt]$\pm$ 0.2} & \shortstack[c]{\textit{56.8}\\[-1pt]$\pm$ 0.4} & \shortstack[c]{\textit{55.3}\\[-1pt]$\pm$ 0.5} \\
		\MINDFull & \shortstack[c]{0.642\\[-1pt]$\pm$ 0.000} & \shortstack[c]{0.492\\[-1pt]$\pm$ 0.009} & \shortstack[c]{0.280\\[-1pt]$\pm$ 0.007} & \shortstack[c]{0.211\\[-1pt]$\pm$ 0.017} & \shortstack[c]{0.019\\[-1pt]$\pm$ 0.075} & \shortstack[c]{\textbf{67.8}\\[-1pt]$\pm$ 0.1} & \shortstack[c]{\textit{64.4}\\[-1pt]$\pm$ 0.4} & \shortstack[c]{\textit{59.1}\\[-1pt]$\pm$ 0.4} & \shortstack[c]{56.2\\[-1pt]$\pm$ 0.8} & \shortstack[c]{54.6\\[-1pt]$\pm$ 0.5} \\
		\MINDCP & \shortstack[c]{\textit{0.643}\\[-1pt]$\pm$ 0.000} & \shortstack[c]{\textbf{0.508}\\[-1pt]$\pm$ 0.010} & \shortstack[c]{\textbf{0.356}\\[-1pt]$\pm$ 0.006} & \shortstack[c]{\textbf{0.315}\\[-1pt]$\pm$ 0.027} & \shortstack[c]{\textbf{0.184}\\[-1pt]$\pm$ 0.076} & \shortstack[c]{\textbf{67.8}\\[-1pt]$\pm$ 0.1} & \shortstack[c]{\textbf{64.6}\\[-1pt]$\pm$ 0.5} & \shortstack[c]{\textbf{59.9}\\[-1pt]$\pm$ 0.3} & \shortstack[c]{\textbf{57.8}\\[-1pt]$\pm$ 0.6} & \shortstack[c]{\textbf{56.3}\\[-1pt]$\pm$ 0.3} \\
		\bottomrule
	\end{tabular}

\end{table*}

\end{document}